\documentclass[10pt,twocolumn,letterpaper]{article}

\usepackage[pagenumbers]{cvpr} %
\usepackage{newtxtext,newtxmath}

\usepackage[accsupp]{axessibility}  %
\usepackage{graphicx}
\usepackage{amsmath}
\usepackage{amssymb}
\usepackage{enumerate}
\usepackage{multirow}
\usepackage{float}
\usepackage[numbers,sort&compress]{natbib}
\usepackage{cite} 
\usepackage{times}
\usepackage{epsfig}
\usepackage{graphicx}
\usepackage{amsmath}
\usepackage{amssymb}
\usepackage{enumerate}
\usepackage{multirow}
\usepackage{xcolor}
\usepackage[normalem]{ulem}
\useunder{\uline}{\ul}{}
\usepackage{wrapfig}
\usepackage{times}
\usepackage{epsfig}
\usepackage{graphicx}
\usepackage{amsmath}
\usepackage{amssymb}
\usepackage{tabulary}
\usepackage{multirow}
\usepackage{silence}
\usepackage{caption}
\usepackage{xfrac}
\usepackage{float}
\usepackage[utf8]{inputenc}
\usepackage[T1]{fontenc}
\usepackage[pagebackref,breaklinks,colorlinks,allcolors=cvprblue]{hyperref}

\renewcommand{\paragraph}[1]{\vspace{0em}\noindent\textbf{#1}.}

\usepackage{color}
\usepackage{xcolor}
\definecolor{turquoise}{cmyk}{0.65,0,0.1,0.3}
\definecolor{purple}{rgb}{0.65,0,0.65}
\definecolor{dark_green}{rgb}{0, 0.5, 0}
\definecolor{orange}{rgb}{0.8, 0.6, 0.2}
\definecolor{red}{rgb}{0.8, 0.2, 0.2}
\definecolor{darkred}{rgb}{0.6, 0.1, 0.05}
\definecolor{blueish}{rgb}{0.0, 0.3, .6}
\definecolor{light_gray}{rgb}{0.7, 0.7, .7}
\definecolor{pink}{rgb}{1, 0, 1}
\definecolor{greyblue}{rgb}{0.25, 0.25, 1}
\definecolor{tab_blue}{HTML}{1f77b4}
\definecolor{tab_orange}{HTML}{ff7f0e}
\definecolor{LightRed}{rgb}{0.99,0.89,0.89}

\definecolor{mesh_misty_rose}{HTML}{e6aaa3}
\definecolor{mesh_yellow}{HTML}{ffba00}

\definecolor{our_red}{rgb}{0.99,0.89,0.89}
\definecolor{our_blue}{HTML}{1f77b4}
\definecolor{our_orange}{HTML}{ff7f0e}

\usepackage{amsmath}

\DeclareMathOperator*{\argmin}{arg\,min}

\makeatletter
\let\old@accent\accent
\DeclareRobustCommand{\accent}[2]{%
  \ifmmode
    \mathaccent #1 #2%
  \else
    \old@accent #1 #2%
  \fi
}
\makeatother

\definecolor{cvprblue}{rgb}{0.21,0.49,0.74}

\usepackage{xspace}
\newcommand{\Ours}{{\textit{MoSca}}\xspace}
\newcommand{\Motion}{{\textit{MoSca}}\xspace}
\newcommand{\mb}[1]{\mathbf{#1}} 
\newcommand{\mc}[1]{\mathcal{#1}} 

\def\paperID{75} %
\def\confName{CVPR}
\def\confYear{2025}

\title{\textit{\textcolor{Fuchsia}{Mo}\textcolor{Dandelion}{Sca}}: Dynamic Gaussian Fusion from Casual Videos via 4D \textcolor{Fuchsia}{Mo}tion \textcolor{Dandelion}{Sca}ffolds}

\author{
        Jiahui Lei\textsuperscript{1} \quad Yijia Weng\textsuperscript{2} \quad  Adam W. Harley\textsuperscript{2} \quad Leonidas Guibas\textsuperscript{2} \quad Kostas Daniilidis\textsuperscript{1,3}\\
        $^1$ University of Pennsylvania \qquad
        $^2$ Stanford University \qquad $^3$ Archimedes, Athena RC\\
        {\tt\small \{leijh, kostas\}@cis.upenn.edu, \{yijiaw, aharley, guibas\}@cs.stanford.edu} 
    }

\begin{document}

\twocolumn[\maketitle\vspace{-3em}
\begin{center}
\includegraphics[width=1.0\linewidth]{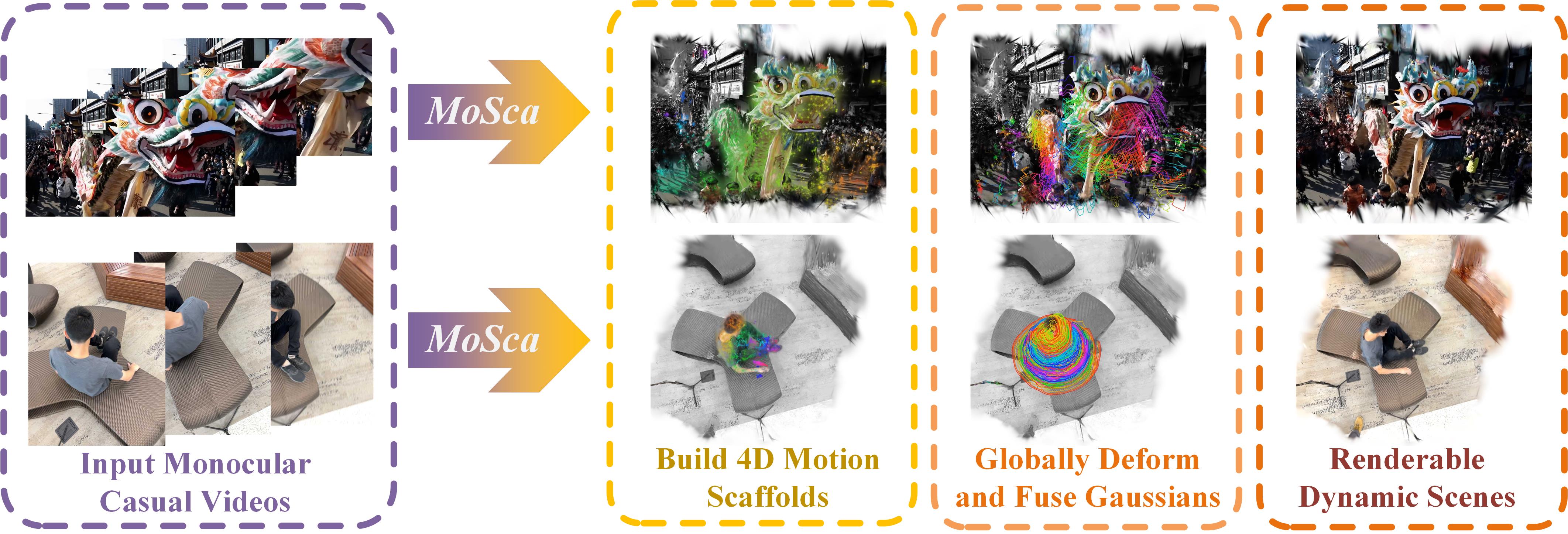}
\captionof{figure}{MoSca reconstructs renderable dynamic scenes from monocular casual videos.}
\end{center}
\label{fig:teaser}
\vspace{-1em}\bigbreak]
\begin{abstract}
We introduce 4D Motion Scaffolds (MoSca), a modern 4D reconstruction system designed to reconstruct and synthesize novel views of dynamic scenes from monocular videos captured casually in the wild. 
To address such a challenging and ill-posed inverse problem, we leverage prior knowledge from foundational vision models and lift the video data to a novel Motion Scaffold (MoSca) representation, which compactly and smoothly encodes the underlying motions/deformations. 
The scene geometry and appearance are then disentangled from the deformation field and are encoded
by globally fusing the Gaussians anchored onto the MoSca and optimized via Gaussian Splatting. 
Additionally, camera focal length and poses can be solved using bundle adjustment without the need of any other pose estimation tools.
Experiments demonstrate state-of-the-art performance on dynamic rendering benchmarks and its effectiveness on real videos.
Project page and code: \url{https://www.cis.upenn.edu/~leijh/projects/mosca}
\end{abstract}

\section{Introduction}
This paper presents 4D Motion Scaffolds (\Ours), a fully automated system for reconstructing and rendering dynamic scenes from casual monocular video inputs with unknown camera parameters---the most typical data format for such a system in the wild. 
Robust 4D scene reconstruction from such input is increasingly vital for constructing datasets for future AGI models, content creation for spatial computing and VR/MR/AR, and building embodied agents to perceive and learn from real video data. 
However, this task is known to be highly challenging and inherently ill-posed~\cite{dycheck, rodynrf, lee2023fast} due to the limited availability of multi-view stereo cues in casual video footage.

To tackle this challenging task, our first insight is to leverage the recent advances of pretrained vision models (Sec.~\ref{sec:method_foundation}), which today are very effective at fundamental
computer vision tasks such as tracking and depth estimation. 
While this knowledge provides a critical boost to understanding the complete dynamic scene, it is inherently insufficient, as it fails to capture occluded parts of the scene and it is usually noisy, local, and partial.
Our second insight is to design a deformation representation, \Ours, derived from the above foundational priors, exploiting a \textit{physical} deformation prior. Although the real-world geometry and appearance are complex and include high-frequency details, the underlying deformation that drives these geometries is usually compact (low-rank) and smooth. \Ours leverages this property by disentangling the 3D geometry and motion, representing the deformation with sparse graph nodes that can be smoothly interpolated (Sec.~\ref{sec:method_mosca}). Another physical prior we exploit is the as-rigid-as-possible (ARAP) deformation, which can be efficiently applied via the trajectory topology of \Ours.
Two important benefits arise from the above two insights: firstly, \Ours can be lifted into 3D and optimized from the inferred 2D foundational priors (Sec.~\ref{sec:method_4dlift}), and secondly, the observations from all timesteps can be globally fused and rendered for any query time (Sec.~\ref{sec:method_scene}).
Gaussian fusion happens when we deform all Gaussians observed at different times to the query time, forming a complete reconstruction, which can be supervised through Gaussian Splatting~\cite{kerbl20233d}.
Furthermore, our system estimates the camera poses and focal lengths via a bundle adjustment and the photometric optimization (Sec.~\ref{sec:method_camera}), obviating the need for other poes estimators such as COLMAP.

\noindent In summary, our main contributions can be summarized as:
(1) An automatic 4D reconstruction system that works in the real world for pose-free monocular videos.
(2) A novel Motion Scaffold deformation representation, which we build using knowledge from 2D foundational models, and optimize via physically-inspired deformation regularization.
(3) An efficient and explicit Gaussian-based dynamic scene representation, driven by \Ours, which globally fuses observations across an input video to render this data into any new viewpoint and query time of choice.
(4) State-of-the-art performance on dynamic scene rendering benchmarks.
\section{Related Works}
\paragraph{Dynamic Novel-View Synthesis}
Novel-view synthesis of dynamic scenes is challenging. Many existing works~\cite{zitnick2004high,stich2008view,bemana2020xfields,bansal20204d,li2022neural3d,pumarola2021d,fridovich2023k,cao2023hexplane,attal2023hyperreel,luiten2023dynamic,lin2023im4d} assume available synchronized multi-view video inputs. 
Another line of works~\cite{yoon2020novel,li2021neural,xian2021space,wang2021neural,gao2021dynamic,tretschk2021non,wu2022d,song2023nerfplayer,tian2023mononerf,you2023decoupling,yang2023deformable,liang2023gaufre,rodynrf,bui2023dyblurf,miao2024ctnerf} tackles the more practical setting of monocular inputs, 
where ambiguities from limited observations further complicate the problem.
As~\cite{dycheck} pointed out, most methods struggle with realistic single-view videos. %
To disambiguate, some works~\cite{athar2022rignerf,weng2022humannerf, rivero2024rig3dgs,liu2024gva,shao2024splattingavatar,svitov2024haha,wen2024gomavatar,lei2023gart,kocabas2023hugs,hu2023gauhuman,hu2023gaussianavatar,chen2023monogaussianavatar,qian20233dgs,li2023human101} target specific scenes and exploit domain knowledge like template models~\cite{bogo2016keep,tran2018nonlinear}. 
A few recent works \cite{li2023dynibar,lee2023fast,zhou2024dynpoint,zhao2024pseudogeneralized} 
fuse information across frames, but only from a small temporal window. %

Neural radiance fields~\cite{mildenhall2021nerf,barron2021mip,chen2022tensorf,muller2022instant,park2021nerfies,park2021hypernerf,fang2022fast} and 3D Gaussian Splatting~\cite{kerbl20233d,gof,keselman2022approximate,keselman2023flexible} are promising approaches to novel view synthesis. The latter's explicit point-based representation fits particularly well into the dynamic setting~\cite{luiten2023dynamic,wu20234d,yang2023deformable,duisterhof2023md,liang2023gaufre,yang2023real,katsumata2023efficient,lin2023gaussian,li2023spacetime,kratimenos2023dynmf,huang2023sc,das2023neural,duan20244d,chu2024dreamscene4d}. We employ 3D Gaussians for long-term, global aggregation. 
Compared to concurrent works~\cite{som2024,stearns2024dgmarbles,liu2024modgs,seidenschwarz2024dynomo}, \Ours has a more structured deformation representation exploiting powerful 2D foundation models, and is a full-stack automated system that directly outputs 4D reconstruction from an unposed RGB video.

\paragraph{Non-Rigid Structure-from-Motion}
Geometric reconstruction of non-rigidly deforming scenes from a single camera is a long-standing problem. \cite{blanz1999morphable,ramakrishna2012reconstructing,zollhofer2014real,bogo2016keep,yang2021lasr,yang2022banmo} focus on specific object categories or articulated shapes and register observations to template models~\cite{bogo2016keep}. 
~\cite{curless1996volumetric,li2008global,dou20153d,newcombe2015dynamicfusion,gao2018surfelwarp,bozic2020neural,du2021neural} warp, align, and fuse scans of generic scenes. To model non-rigid deformations, state-of-the-art methods \cite{dou20153d,zollhofer2014real,newcombe2015dynamicfusion,bozic2020neural} 
use Embedded Deformation Graphs~\cite{sumner2007embedded}, where dense transformations over the space are modeled with a sparse set of basis transformations. 
In ~\Ours, we extend classic Embedded Graphs to connect priors from 2D foundation models to dynamic Gaussian splatting.

\paragraph{2D Vision Foundation Models}
Recent years have witnessed great progress in large-scale pretrained vision foundation models~\cite{bommasani2021opportunities,radford2021learning,dinov2,kirillov2023segment,openai2023gpt4} that serve various downstream tasks, ranging from image-level tasks such as visual question answering~\cite{liu2023improved,liu2024visual,openai2023gpt4} to pixel-level tasks including segmentation~\cite{kirillov2023segment}, dense tracking~\cite{cotracker,pips}, and monocular depth estimation~\cite{zoedepth,unidepth,yang2024depth}. 
These models encode strong data priors 
particularly useful in monocular video-based dynamic reconstruction, as they help disambiguate partial observations. %
While most previous methods~\cite{li2021neural,gao2021dynamic,lee2023fast,zhao2024pseudogeneralized,li2023dynibar,chu2024dreamscene4d,som2024,stearns2024dgmarbles,liu2024modgs} directly use the 2D priors for regularization in image space, and often in isolation from each other, we propose to lift these 2D priors to 3D and fuse them in a coordinated way. %

\section{Method}
\begin{figure*}[t]
    \centering
    \includegraphics[width=1.0\linewidth]{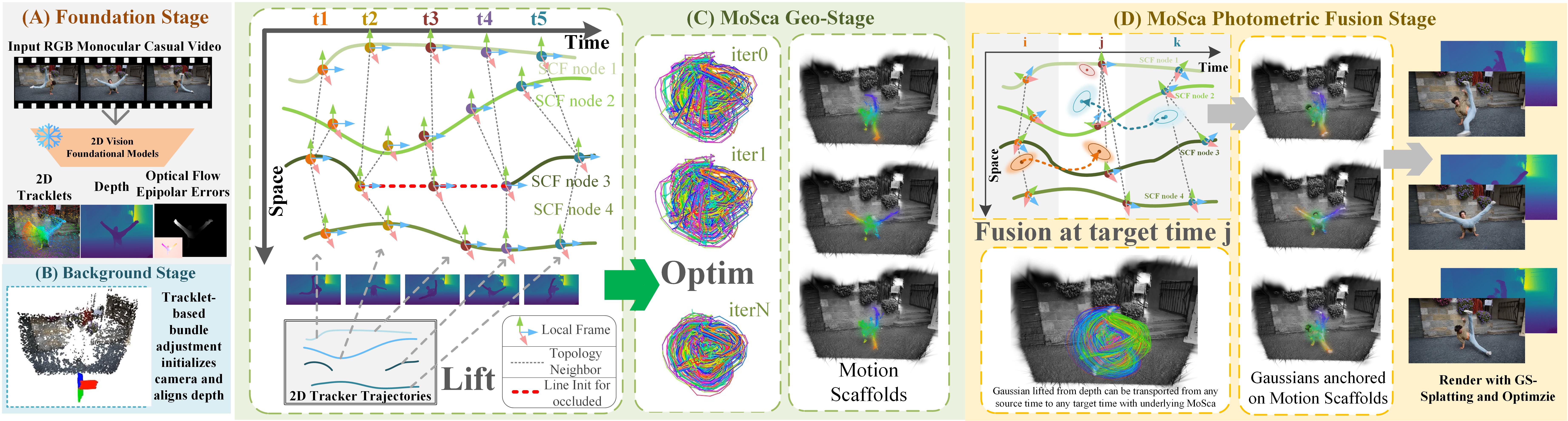}
    \caption{\small\textbf{Overview}: 
    (A) Given a monocular casual video, we infer pre-trained 2D vision foundation models (Sec.~\ref{sec:method_foundation}). 
    (B) The camera intrinsics and poses are initialized using tracklet-based bundle adjustment (Sec.~\ref{sec:method_camera}).
    (C) Our proposed Motion Scaffold (\Motion) is lifted from 2D predictions and optimized with physics-inspired regularizations (Sec.~\ref{sec:method_4dlift}). 
    (D) Gaussians are initialized from all timesteps, deformed with \Motion (Sec.~\ref{sec:method_mosca}), and fused globally to model the dynamic scene. The entire representation is rendered with Gaussian Splatting and optimized with photometric losses (Sec.~\ref{sec:method_scene}).
    }
    \label{fig:main}
    \vspace{-1.5em}
\end{figure*}

\paragraph{Overview} Given a casual monocular video of a dynamic scene with $T$ frames $\mc{I} = [I_1, I_2, \dots I_T]$, our fully automatic system reconstructs the geometry and appearance of the scene with a set of dynamic Gaussians and recovers the focal length and poses of the camera if they are unknown .%
Our key idea is to lift the 2D video input to a novel 4D dynamic scene representation, which we name Motion Scaffolds (\Ours), where all the observations are fused \textbf{globally} and \textbf{geometrically}. 
Fig.~\ref{fig:main} provides an overview of our approach.
We first introduce the deformation representation \Ours in Sec.\ref{sec:method_mosca} and then, detail each step of our reconstruction system in Sec.~\ref{sec:method_system}.

\subsection{Deformation Representation with \Ours}
\label{sec:method_mosca}
A fundamental challenge in real-world 4D reconstruction is the high dimensionality of the potential solution space compared to the extremely limited spatiotemporal observations. 
However, real-world motion typically behaves rigidly, smoothly, and compactly, meaning that the actual solution is low-rank and driven by a few key ``eigen'' motions. With this insight, we model the underlying deformation of the scene using an explicit, compact, and structured graph $(\mathcal{V}, \mathcal{E})$, named 4D Motion Scaffold (\Motion), which encodes these local ``eigen'' motions and interpolates the dense deformation field.

\paragraph{Motion Scaffold Graph Definition}
Intuitively, the \Ours graph nodes $\mc{V} = \{\mb v^{(m)}\}_{m=1}^{M}$ are 6-DoF trajectories that capture the underlying low-rank, smooth motion of the scene. The number of nodes $M$ is significantly smaller (e.g., see Tab.~\ref{tab:others}) than the number of points required to represent the scene.
Specifically, each node $\mb v^{(m)} \in \mc{V}$ consists of per-timestep rigid transformations $\mb{Q}_t^{(m)}$ and a global control radius $r^{(m)}$, which parameterizes a radial basis function (RBF) describing its influence on nearby space: 
\begin{align}
    \mb v^{(m)} = ([\mb Q^{(m)}_1,\mb  Q^{(m)}_2, \ldots,\mb  Q^{(m)}_T],r^{(m)}),
\end{align}
where $\mb Q^{(m)}=[\mb R^{(m)}, \mb t^{(m)}]\in SE(3)$ and $r^{(m)}\in \mathbb{R}^+$ is the radius.
To properly interpolate the node-encoded trajectories and regularize the deformation, we organize the nodes $\mb v^{(m)}$ into a topology. We define the \Ours graph edges $\mc{E}$ as:
\vspace{-1em}
\begin{align}
    &\mc{E}(m) = \text{KNN}_{n\in \{1,\ldots, M\}}\left[D_{\text{curve}}(m,n)\right], \nonumber \\
    & D_{\text{curve}}(m,n) = \max_{t=1,2,\ldots, T} \|\mb t^{(m)}_t - \mb t^{(n)}_t\|,
    \label{eq: def_topo}
\end{align}
where $\text{KNN}$ denotes the K-nearest neighbors under the curve distance metric $D_{\text{curve}}$. This metric captures the global proximity between trajectories across all timesteps and accounts for topological changes (e.g., opening a door does not connect the door and wall).

\paragraph{$\mb{SE(3)}$ Deformation Field}
Given \Motion $(\mc V, \mc E)$, we can derive a dense deformation field by interpolating motions from nodes near the query point. We use Dual Quaternion Blending (DQB)~\cite{DQB} to mix multiple $SE(3)$ elements on the $SE(3)$ manifold.
Similar to the unit quaternion representation of $SO(3)$, the unit dual quaternion represents $SE(3)$ using eight numbers by including a dual part. Please refer to~\cite{jia2013dual, DQB, daniilidis1999hand} for details. Given $L$ rigid transformations $\mathbf{Q}_i \in SE(3)$ and their blending weights $w_i$, the interpolated motion is: 
\vspace{-1em}
\begin{equation}
   \text{DQB}(\{(w_i, \mb Q_i)\}_{i=1}^L) = \frac{\sum_{i=1}^Lw_i \hat{\mb q}_i}{\|\sum_{i=1}^Lw_i \hat{\mb q}_i\|_{DQ}} \in SE(3) ,
\end{equation}
where $\hat{\mathbf{q}}$ is the dual quaternion representation of $\mathbf{Q}$ and $|\cdot|_{DQ}$ denotes the dual norm~\cite{DQB}. Unlike linear blend skinning (LBS), DQB is a manifold interpolation that always produces an interpolated element in $SE(3)$.
Consider any query position $\mb x$ in 3D space at time $t_{\text{src}}$. Denote its nearest node at $t_{\text{src}}$ as $\mb v^{(m^*)}$ where $m^* = \argmin_{m} ||\mb t_{t_{src}}^{(m)} - \mb x||$ and $\mb t_{t_{src}}^{(m)}$ is the translation part of node $m$'s transformation at time $t_{src}$. 

We can efficiently compute its $SE(3)$ deformation to the query time $t_{\text{dst}}$ using nodes in the neighborhood of $v^{(m^*)}$. Formally, the deformation field $\mc W$ from time $t_{\text{src}}$ to time $t_{\text{dst}}$ is: 
\vspace{-0.5em}
\begin{align}
    \mc W (\mb x, \mb w; t_{\text{src}}, t_{\text{dst}}) = \text{DQB}\left(\{w_i, \Delta \mb Q^{(i)}  \}_{i\in \mc{E}(m^*)}\right), 
    \label{eq: warping}
\end{align}
where $\Delta \mb Q^{(i)}=\mb Q^{(i)}_{t_{\text{dst}}} \ (\mb Q^{(i)}_{t_{\text{src}}})^{-1}$ and $\mb w = \{w_i\}$ are skinning weights computed from RBFs parameterized by radius $r^{(i)}$: 
\vspace{-1em}
\begin{equation}
    w_i(\mb x, t_{\text{src}}) =\exp{(-{\|\mb x-\mb t^{(i)}_{t_{\text{src}}}\|_2^2}/{2r^{(i)}})} \in \mathbb R^+
    \label{eq: skinning}.
\end{equation}
In summary, \Ours $(\mc{V}, \mc{E})$ encodes the deformation field through skinning on a structured, sparse trajectory graph. In the following sections, we will demonstrate how to reconstruct \Ours and attach Gaussians onto it to produce the final 4D reconstruction.

\subsection{Reconstruction System}
\label{sec:method_system}
\subsubsection{Leveraging Priors from 2D Foundation Models}
\label{sec:method_foundation}
\vspace{-0.5em}
4D reconstruction from monocular videos is highly ill-posed; therefore, it is essential to leverage prior knowledge to constrain the solution space. In the first step of our system, we exploit the priors provided by large vision foundation models pre-trained on massive datasets. Specifically, we utilize off-the-shelf pre-trained models to obtain:
1) Depth estimations~\cite{unidepth,metric3dv2,depthcrafter} $\mc{D} = [D_1, D_2, \ldots, D_T]$ that are relatively consistent across frames; 
2) Long-term 2D pixel trajectories~\cite{doersch2024bootstap,karaev2024cotracker3,SpatialTracker} $\mc{T}=\{\tau^{(i)}=[(p^{(i)}_1,v^{(i)}_1), (p^{(i)}_2, v^{(i)}_2),\ldots, (p^{(i)}_T, v^{(i)}_T)]\}_i$, where $p^{(i)}_t$ and $v^{(i)}_t$ represent the $i$-th trajectory's 2D image coordinate and visibility at frame $t$;
3) Per-frame epipolar error maps $\mc{M}=[E_1, E_2, \ldots, E_T]$~\cite{rodynrf} computed from RAFT\cite{raft} dense optical flow predictions, which indicate the likelihood of being in the dynamic foreground.
These inferred results provide critical cues about geometry and correspondence. However, such raw information is partial, local, and noisy, and does not constitute a complete solution. We are going to fuse and optimize these initial cues to produce a coherent and global 4D reconstruction.

\vspace{-0.5em}
\subsubsection{Camera Initializaition}
\label{sec:method_camera}
\vspace{-0.5em}
To enable 4D reconstruction in the wild, our system must operate on dynamic scene videos with unknown camera parameters. Therefore, in the second step of our reconstruction pipeline, we propose a tracklet-based bundle adjustment to robustly initialize the camera focal lengths and poses.
Given the inferred 2D tracks $\mathcal{T}$ and epipolar error maps $\mathcal{M}$, we first compute the maximum epipolar error of each tracklet as $e(\tau) = \max_{t=1\ldots T} E_t[p_t] \cdot v_t$ across visible timesteps. We identify confident background tracklets by thresholding $e(\tau)$ with a predefined small threshold. Starting with a pre-defined initial camera focal length, we optimize the camera poses and intrinsics jointly by minimizing the reprojection errors on these confident static tracks:
\vspace{-1em}
\begin{align}
    & \mc L_{proj} = \sum_{i\in |\mc T_{\text{static}}|} \sum_{a,b \in [1,T]} (v^{(i)}_{a}v^{(i)}_{b})  \\& \cdot  \left\| \pi_{\mb K}\left(\mb W^{-1}_{b} \mb W_{a} \pi_{\mb K}^{-1}(p^{(i)}_{a}, D_{a}[p^{(i)}_{a}])\right) - p^{(i)}_b \right \|, \nonumber
    \label{eq: loss_ba_proj}
\end{align}
where $p_a$ and $p_b$ are pixel locations, $\pi_{\mathbf{K}}$ denotes projection with intrinsics $\mathbf{K}$, and $\mathbf{W}_{t}$ is the camera pose at time $t$.
To account for errors in the depth estimation—particularly scale misalignment—we jointly optimize a correction to the depth $D_{a}[p_{a}]$, which consists of per-frame global scaling factors and small per-pixel corrections, using a depth alignment loss:
\vspace{-1.0em}
\begin{align}
    & \mc L_{z} = \sum_{i\in |\mc T_{\text{static}}|} \sum_{a,b \in [1,T]} (v^{(i)}_{a}v^{(i)}_{b}) \\ &D_\text{scale-inv}
    \left( 
        \left[\mb W^{-1}_{b} \mb W_{a} \pi_{\mb K}^{-1}(p^{(i)}_{a}, D_{a}[p^{(i)}_{a}])\right]_z,  D_{b}[p^{(i)}_b]
    \right),  \nonumber
    \label{eq: loss_ba_dep}
    \vspace{-1em}
\end{align}
where $[\cdot]_z$ takes the $z$ coordinate, and $D_\text{scale-inv}(x,y) = |x/y -1| + |y/x - 1|$. 
The overall bundle adjustment loss is $\mc L_{\text{BA}} = \lambda_\text{proj} \mc L_{proj} + \lambda_\text{z} \mc L_{z}$, and the solved camera poses $\mathbf{W}_t$ will be refined during later rendering phases. While camera solving is not our primary contribution, our system achieves state-of-the-art camera pose accuracy on dynamic videos (Sec.~\ref{sec:exp_camera}); more details are provided in the Supplemental Material.
\begin{figure*}[t]
    \centering
    \includegraphics[width=1.0\linewidth]{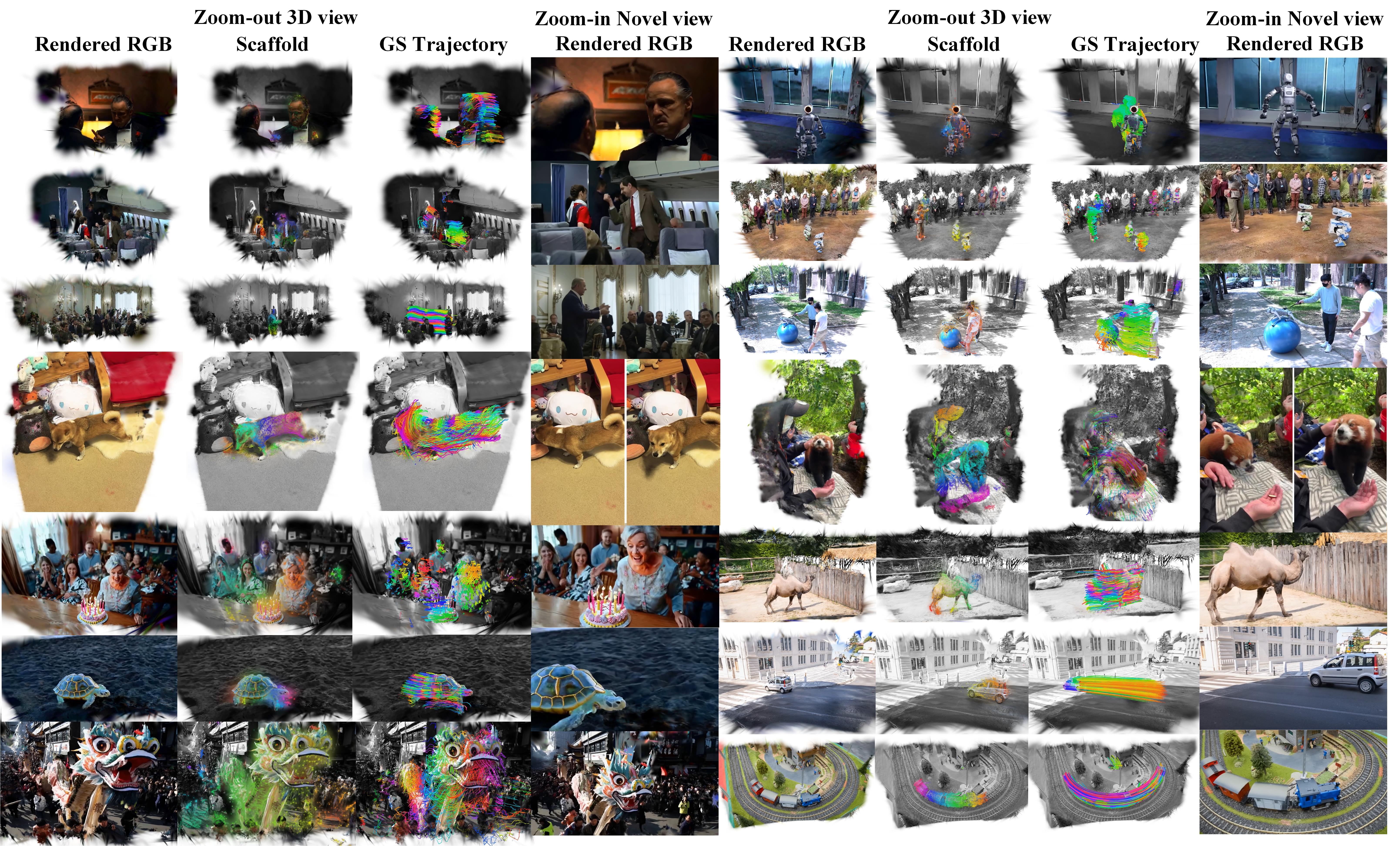}
    \caption{\textbf{In-the-wild videos}: \Ours can process a list of RGB frames and reconstruct the 4D scene from various types of videos.}
    \label{fig:wild}
    \vspace{-1em}
\end{figure*}

\vspace{-1em}
\subsubsection{Geometric Optimization of \Ours}
\label{sec:method_4dlift}
After inferring the 2D foundational models and initializing the camera, we are ready to geometrically construct \Ours $(\mathcal{V}, \mathcal{E})$ in the third step of our system. 
A key contribution of this paper is the seamless integration of \Ours with powerful 2D foundational models. Specifically, the long-term 2D tracking $\mathcal{T}$, together with the depth estimates $\mathcal{D}$, provide strong cues for constructing $\mathcal{V}$. However, there is still a gap due to missing information when tracks are invisible and because the local rotation component of \Ours is also unknown. We address this gap by incorporating physics-inspired regularization into the optimization of \Ours.

\paragraph{3D Lift and Initialization} Similar to the camera initialization, we identify foreground 2D tracks by thresholding the maximum epipolar error $e(\tau)$ of each tracklet. We then lift the foreground tracklets into 3D using depth estimates $\mathcal{D}$ at visible timesteps and linearly interpolate between nearby observations at occluded timesteps. Formally, we compute the lifted 3D position $\mathbf{h}_t$ at timestep $t$ from the 2D track $\tau = [(p_t, v_t)]_{t=1}^T$ as 
\begin{equation}
    \mb h_t = \begin{cases}
        & \mb W_t \pi^{-1}_{\mb K}(p_t, D_t[p_t]), \quad \text{if} \quad v_t=1, \\
        & \text{LinearInterp}(\mb h_{\text{left}}, \mb h_{\text{right}}), \quad \text{if} \quad v_t=0,
    \end{cases}
    \label{eq: lift}
\end{equation}
where $\pi^{-1}_{\mathbf{K}}$ refers to back-projection with camera intrinsics $\mathbf{K}$, $\mathbf{W}_t$ refers to the camera pose, and $\mathbf{h}_{\text{left}}, \mathbf{h}_{\text{right}}$ refer to the lifted 3D positions from the nearest visible timesteps before and after $t$.
From each track, we initialize a \Motion node $\mathbf{v}^{(i)}$ using the lifted positions $\mathbf{h}_t$ as the translation part and the identity as the rotation, i.e., $\mb Q^{(i)}_t = [\mb I, \mb h_t^{(i)}]$, along with a predefined control radius $r_{\text{init}}$. 
In practice, we retain only a subset of the densely inferred 2D tracklets by uniformly resampling nodes based on the curve distance (Eq.~\ref{eq: def_topo}).

\paragraph{Geometry Optimization} Starting from the initialized rotations and the invisible lines, we propagate the visible information to the unknowns through the \Ours topology $\mathcal{E}$ by optimizing a physics-inspired as-rigid-as-possible (ARAP) loss. Given two timesteps separated by a time interval $\Delta$, we define the ARAP loss $\mathcal{L}_{\text{arap}}$ as: 
\begin{align}
    \mc L_{\text{arap}}&= \sum_{t=1}^T \sum_{m=1}^M \sum_{n\in \hat {\mc E}(m)}  \lambda_{\text{l}}\left | \| \mb t_t^{(m)} - \mb t_t^{(n)}\| - \| \mb t_{t+\Delta}^{(m)} - \mb t_{t+\Delta}^{(n)}\| \right | \nonumber
    \\
    &+ \lambda_{\text{c}} \left\| \mb Q^{-1 \, (n)}_t \mb t^{(m)}_t - \mb Q^{-1 \, (n)}_{t+\Delta} \mb t^{(m)}_{t+\Delta} \right\|,
    \label{eq: loss_arap}
\end{align}
where $\hat {\mathcal{E}}$ refers to a multi-level sub-sampled topology pyramid from $\mathcal{E}$ in \Ours (detailed in the Supplemental Material). The first term encourages the preservation of local distances in the neighborhood, and the second term preserves the local coordinates by involving the local frame $\mathbf{Q}$ in the optimization.
We also enforce the temporal smoothness of the deformation by regularizing the velocity and acceleration: 
\vspace{-1em}
\begin{align}
    \mc L_\text{vel} &= \sum_{t=1}^T\sum_{m=1}^M \|\mb t_t^{(m)} - \mb t_{t+1}^{(m)}\| + \|\log(\mb R_{t}^{(m)} \mb R_{t+1}^{-1\,(m)})\|_F \nonumber \\ 
    \mc L_\text{acc} &= \sum_{t=1}^T\sum_{m=1}^M \|\mb t_t^{(m)} - 2\mb t_{t+1}^{(m)} + t_{t+2}^{(m)}\|  \\ &+ \left| \|\log(\mb R_{t}^{(m)} \mb R_{t+1}^{-1\,(m)})\|_F -\|\log(\mb R_{t+1}^{(m)} \mb R_{t+2}^{-1\,(m)})\|_F \right|, \nonumber
    \label{eq: loss_vel_acc}
\end{align}
where $\|\log(\cdot)\|_F$ refers to the Frobenius norm of rotation logarithm (the axis-angle of the rotation).
In summary, the objective of this geometric optimization in the third step of our system is $\mathcal{L}_{\text{geo}} = \lambda_{\text{arap}} \mathcal{L}_\text{arap} + \lambda_{\text{acc}} \mathcal{L}_\text{acc} + \lambda_{\text{vel}} \mathcal{L}_\text{vel}$, and we only optimize rotations and invisible 3D translations, leaving the visible 3D positions unchanged to prevent degeneration.

\subsubsection{Photometric Optimization of \Ours}
\label{sec:method_scene}
\begin{figure*}[t]
    \centering
    \includegraphics[width=1.0\linewidth]{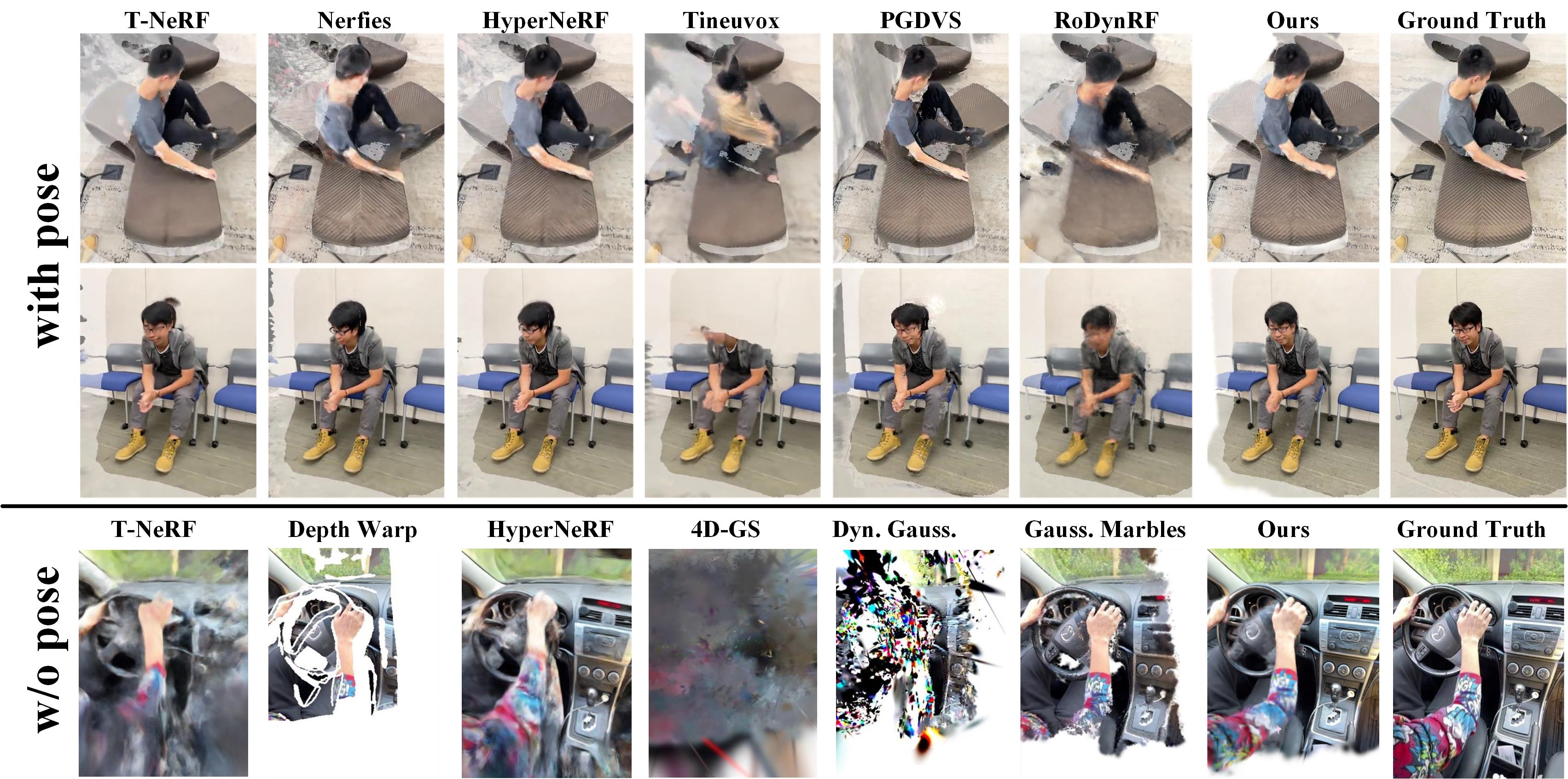}
    \caption{
    Visual comparison on DyCheck~\cite{dycheck} under the settings with or without camera pose.
    }
    \label{fig:dycheck}
    \vspace{-1em}
\end{figure*}

\paragraph{Dynamic Scene Representation}
An important feature of \Ours is that its global deformation field can transform points at any time globally, enabling the fusion of all observed video frames into a single coherent representation. In the final step of the system, the optimized \Ours collects 3D Gaussians initialized from back-projected foreground depth points at all timesteps. Formally: 
\begin{equation}
    \mc G = \{(\mu_j, R_j, s_j, o_j, c_j; t^{\text{ref}}_j, \Delta \mb w_j) \}_{j=1}^N,
    \label{eq: dyn_gs}
\end{equation}
where the first five attributes are the center, rotation, non-isotropic scales, opacity, and spherical harmonics of 3DGS~\cite{kerbl20233d}, and the latter two are tailored for \Ours. Specifically, $t^{\text{ref}}_j$ is the reference timestep—that is, the timestep at which the Gaussian is initialized from the back-projected depth; and $\Delta \mathbf{w}_j \in \mathbb{R}^K$ is the per-Gaussian learnable skinning weight correction.
To obtain the complete geometry at a query timestep $t$, Gaussians from all timesteps are deformed to the query time $t$ and fused: 
\begin{align}
    \mc G(t) &= \{(\mb T_j(t) \mu_j, \mb T_j(t) R_j, s_j, o_j, c_j)\, |\, \nonumber \\ &\mb T_j(t) = \mc W (\mu_j, \mb w (\mu_j, t^{\text{ref}}_j) + \Delta \mb w_j; t^{\text{ref}}_j, t) \}_{j=1}^N
    \label{eq: final_repr}
\end{align}
where $\mathcal{W}$ is the deformation field defined in Eq.\ref{eq: warping}, and $\mathbf{w}$ is the base RBF skinning weight defined in Eq.\ref{eq: skinning}.
The static background is also represented as a standard 3DGS $\mathcal{H} = {(\mu_j, R_j, s_j, o_j, c_j) }_{j=1}^H$, which can be initialized by back-projecting the depth map using known camera parameters. Therefore, the final renderable dynamic scene at time $t$ can be approximated by the union $\mathcal{G}(t) \cup \mathcal{H}$.

\paragraph{Photometric Optimization}
The Gaussians described above can be rendered using a Gaussian Splatting-based differentiable renderer and optimized with depth and RGB rendering losses, along with the regularization losses from Sec.~\ref{sec:method_4dlift}. To fully exploit the inferred tracklets, we also render a flow/track map by rasterizing the XYZ coordinates (replacing the RGB color with XYZ values) of each Gaussian at different timesteps. We supervise the flow/track map with the inferred 2D tracklets as a regularization loss $\mathcal{L}_\text{track}$~\cite{som2024}. The final photometric step has a total objective: 
\vspace{-1em}
\begin{align}
    \mc L &= \lambda_\text{rgb}\mc L_\text{rgb} + \lambda_\text{dep}\mc L_\text{dep} + \lambda_\text{track}\mc L_\text{track} \nonumber
    \\ &+ \lambda_\text{arap}\mc L_\text{arap} + \lambda_\text{acc}\mc L_\text{acc}+ \lambda_\text{vel}\mc L_\text{vel}.
\end{align}

\paragraph{Node Control}
Similar to standard 3DGS Gaussian control techniques including gradient-based densification and reset-pruning simplification, we propose a novel control policy over the proposed \Motion nodes. 
To periodically densify nodes, we select Gaussians with high tracking-loss $\mathcal{L}_\text{track}$ induced gradients, subsample them, and convert them into new \Ours nodes. 
To clean the representation and prune the structure, we also periodically copy the dynamic foreground Gaussians from a randomly selected timestep into the static background and reset the foreground Gaussians to a low opacity. This simplifies unnecessary foreground Gaussians. We then prune nodes whose skinning weights toward all Gaussians fall below a threshold, indicating a limited contribution to deformation modeling.

\section{Experiments}
\subsection{Novel View Synthesis}
\label{sec:exp_rendering}
\paragraph{In-the-wild}
One of the most significant results of \Ours is demonstrating that such an automatic dynamic rendering system can work effectively in real-world scenarios. In Fig.~\ref{fig:wild}, we showcase reconstruction results on diverse in-the-wild \textbf{monocular} videos—including movie clips, internet videos, SORA-generated videos, and DAVIS\cite{davis} videos—demonstrating the effectiveness of \Ours.
\begin{table}[t]
\caption{Comparison on DyCheck~\cite{dycheck}, group w-pose and w/o-pose means with or without camera pose and are averaged over all 7 scenes on the standard 2x resolution. Group SOM-5-1x means using the 5 scenes and 1x res. as in Shape-of-Motion~\cite{som2024}.}
\scalebox{0.77}{
\begin{tabular}{c|c|ccc}
\hline
                                                  & { Method}                   & mPSNR$\uparrow$ & mSSIM$\uparrow$ & mLPIPS$\downarrow$ \\ \hline
                           & T-NeRF~\cite{dycheck}                           & 16.96 & 0.577 & 0.379  \\
                           & NSFF~\cite{li2021neural}                                            & 15.46 & 0.551 & 0.396  \\
                           & Nerfies~\cite{park2021nerfies}                                         & 16.45 & 0.570 & 0.339  \\
                           & HyperNeRF~\cite{park2021hypernerf}                                       & 16.81 & 0.569 & 0.332  \\
                           & PGDVS~\cite{zhao2024pseudogeneralized}                                           & 15.88 & 0.548 & 0.340  \\
                           & DyPoint~\cite{zhou2024dynpoint}                                         & 16.89 & 0.573 & -      \\
                           & DpDy~\cite{wang2024diffusion}                                            & -     & 0.559 & 0.516  \\
                           & Dyn.Gauss.~\cite{luiten2023dynamic}                                  & 7.29  & -     & 0.692  \\
                           & 4D GS~\cite{wu20234d}                                           & 13.64 & -     & 0.428  \\
                           & Gauss.Marbles~\cite{stearns2024dgmarbles}                                & 16.72 & -     & 0.413  \\
                           & DyBluRF~\cite{bui2023dyblurf}                                         & 17.37 & 0.591 & 0.373  \\
                           & CTNeRF~\cite{miao2024ctnerf}                                          & 17.69 & 0.531 & -      \\
                           & D-NPC~\cite{kappel2024d}                                           & 16.41 & 0.582 & 0.319  \\ 
                           & Shape-of-Motion~\cite{som2024}                                           & 17.32 & 0.598 & 0.296  \\ 
\multirow{-14}{*}{{w-pose}}  & \textbf{Ours}                                            & \textbf{19.32} & \textbf{0.706} & \textbf{0.264}  \\ \hline
                           & RobustDynrf~\cite{rodynrf}                                     & 17.10 & 0.534 & 0.517  \\
                           & { Dyn.Gaussians}~\cite{luiten2023dynamic}           & 7.60  & -     & 0.704  \\
                           & { 4D GS}~\cite{wu20234d}                    & 13.11 & -     & 0.726  \\
                           & { Gaussian Marbles}~\cite{stearns2024dgmarbles}         & 15.79 & -     & 0.430  \\
                           & \textbf{Ours}                                   & 18.84 & 0.676 & 0.289  \\
\multirow{-6}{*}{{w/o-pose}} & { \textbf{Ours} (w. focal)} & \textbf{19.02} & \textbf{0.683} & \textbf{0.279}  \\ \hline

                           & Shape-of-Motion~\cite{som2024}                                     & 16.72 & 0.63 & 0.45  \\
\multirow{-2}{*}{{SOM-5-1x}} & {\textbf{Ours}} & \textbf{18.40} & \textbf{0.67} & \textbf{0.42}  \\ \hline

\end{tabular}
}
\label{tab:iphone}
\vspace{-2em}
\end{table}

\begin{table}[b]
\centering
\caption{Comparison on NVIDIA~\cite{yoon2020novel}, averaged over all scenes. ``w/o'' means without camera pose.}
\scalebox{0.72}{
\begin{tabular}{c|cc|c|cc}
\hline
\multicolumn{1}{c|}{Method} & \multicolumn{1}{c}{PSNR} & \multicolumn{1}{c|}{LPIPS} & \multicolumn{1}{c|}{Method} & PSNR                 & LPIPS                \\ \hline
D-NeRF~\cite{pumarola2021d}                      & 21.49                    & 0.232                      & CTNeRF~\cite{miao2024ctnerf}                      & 26.13       & 0.082                \\
NR-NeRF~\cite{tretschk2021non}                     & 19.69                    & 0.323                      & DynPoint~\cite{zhou2024dynpoint}                    & 26.53                & \uline{0.068}                \\
TiNeuVox~\cite{fang2022fast}                     & 19.74                    & 0.285                      & D-NPC~\cite{kappel2024d}                       & 25.64                & 0.109                \\
HyperNeRF~\cite{park2021hypernerf}                   & 17.60                    & 0.367                      & RoDynRF~\cite{rodynrf}                     & 25.89                & \textbf{0.067}                \\
NSFF~\cite{li2021neural}                        & 24.33                    & 0.199                      & RoDynRF~\cite{rodynrf} w/o             & 25.38                & 0.079                \\
DynNeRF~\cite{gao2021dynamic}                     & 26.10                    & 0.082                      & GaussianMarbles~\cite{stearns2024dgmarbles}            & 22.32                & 0.129                \\
MonoNeRF~\cite{tian2023mononerf}                    & 25.62                    & 0.106                      & \textbf{Ours}                        & \textbf{26.72}                & 0.070                \\
4DGS~\cite{wu20234d}                        & 21.45                    & 0.199                      & \textbf{Ours w/o}                & \uline{26.54}                & 0.073                \\
Casual-FVS~\cite{lee2023fast}                  & 24.57                    & 0.081                      &                             & 
\\ \hline
\multicolumn{1}{l}{} & \multicolumn{1}{l}{}
\end{tabular}
\label{tab:nvidia}
}
\vspace{-2em}
\end{table}

\begin{figure*}[t]
    \centering
    \includegraphics[width=1.0\linewidth]{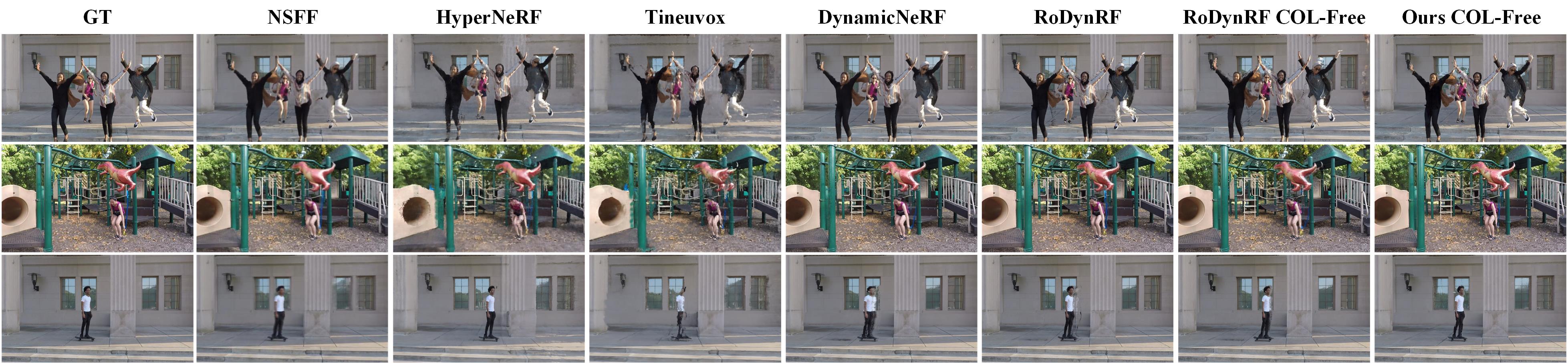}
    \caption{\small Visual comparison on NVIDIA dataset~\cite{yoon2020novel}.}
    \label{fig:nvidia}
    \vspace{-1em}
\end{figure*}
\paragraph{DyCheck}
To quantitatively evaluate our rendering results, we compare our method to others on the currently most challenging dataset -- the iPhone DyCheck~\cite{dycheck}. DyCheck features generic, diverse dynamic scenes captured with a handheld iPhone using realistic camera motions for training, and utilizes two static cameras at significantly different poses from the training views for testing.
For a fair comparison with previous methods that exploit noisy LiDAR depth from the dataset, we use the iPhone's noisy LiDAR depth as the metric depth $\mc D$ and employ BootsTAPIR~\cite{doersch2024bootstap} for tracking. Since the camera parameters are optimized during training, during inference, we fix the scene representation and adjust the test camera poses to find the correct viewpoints.
The quantitative results are reported in Tab.~\ref{tab:iphone}, and qualitative results are shown in Fig.~\ref{tab:iphone}. 
Due to the large deviation of the testing views from the training camera trajectory, most per-frame depth warping methods fail directly (e.g., see Fig.10 of Casual-FVS~\cite{lee2023fast}). Similarly, local fusion methods exhibit large missing areas (e.g., PGDVS~\cite{zhao2024pseudogeneralized}, Gaussian Marbles~\cite{stearns2024dgmarbles}), even though these missing areas are visible in other time steps.
Some recent Gaussian-based methods like 4D-GS~\cite{wu20234d} also fail because they depend on strong multi-view stereo cues to reconstruct the scene. 
As shown in Tab.~\ref{tab:iphone}, we outperform all other methods by a large margin.
We attribute this improvement to two factors: firstly, by leveraging powerful pre-trained 2D long-term trackers, our \Motion representation models long-term motion trajectories, enabling the global aggregation of observations across all timesteps, which leads to a more complete reconstruction. Secondly, the structured sparse motion graph design of \Ours facilitates optimization. Compared to dense Gaussian geometries, its compact and smoothly interpolated motion nodes significantly reduce the optimization space. Its topology enables the effective propagation of information to unobserved regions through ARAP regularization.
Note that our system still performs well under the pose-free setup, as shown in the bottom group of Tab.~\ref{tab:iphone}.

\paragraph{NVIDIA}
We also evaluate \Ours on the widely used NVIDIA video dataset~\cite{yoon2020novel}, following the protocol in RoDynRF~\cite{rodynrf}. As reported in Tab.~\ref{tab:nvidia} and Fig.~\ref{fig:nvidia}, we achieve high PSNR and very competitive LPIPS results.
Since the facing-forward, the small-baseline setting is relatively easier compared to the realistic DyCheck dataset, where most areas of the dynamic scene are visible in neighboring time frames, reducing the need for strong regularization and fusion of information in occluded areas -- the advantages of \Ours are not fully showcased on NVIDIA videos.

\subsection{Camera and Correspondence}
\label{sec:exp_camera}
\paragraph{Camera Pose}
\begin{figure}[b]
    \vspace{-1.5em}
    \centering
    \includegraphics[width=1.0\linewidth]{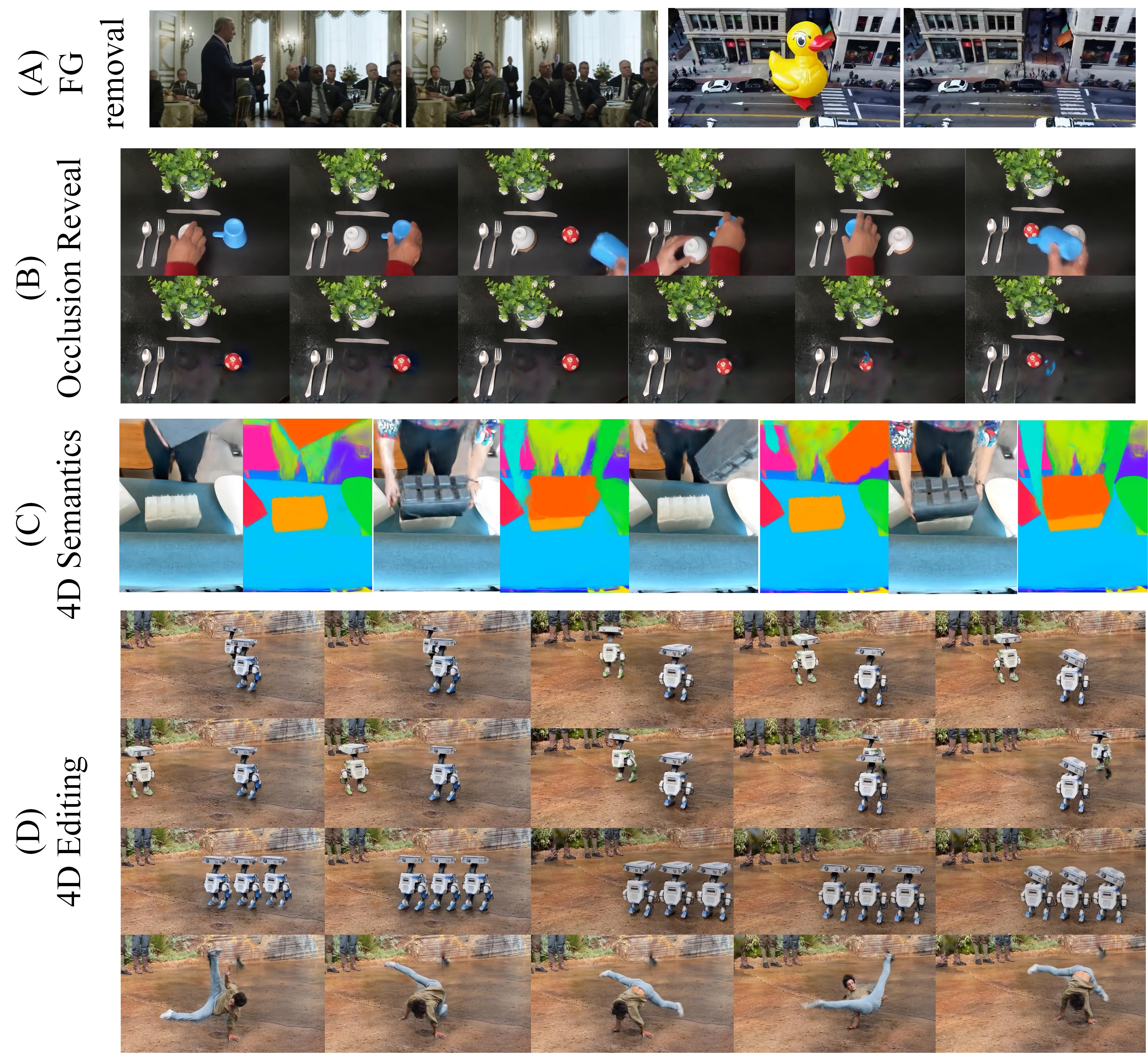}
    \caption{Application of \Ours reconstructed 4D scenes.}
    \label{fig:application}
    \vspace{-2.5em}
\end{figure}
Another advantage of \Ours is its natural integration of camera solving, both geometrically through tracklet-based bundle adjustment and photometrically through rendering-based refinement. We quantitatively evaluate the camera pose estimation, a byproduct of our system, following MonST3R~\cite{zhang2024monst3r} on the SLAM dataset TUM-dynamics~\cite{sturm2012benchmark} and the synthetic Sintel dataset~\cite{butler2012naturalistic}. The camera pose errors are shown in Table~\ref{tab:camera}. Although camera pose estimation is not the main focus of \Ours, it still achieves comparable or even superior performance compared to camera-pose-tailored SLAM-based and DuST3R-based methods. Notably, some of the SLAM systems in the table require known camera intrinsics, whereas \Ours does not.
\begin{table}[t]
\centering
\caption{Camera pose accuracy ($^*$ requires ground truth camera intrinsics as input)}
\scalebox{0.6}{
\begin{tabular}{lccc|ccc}
\hline
                         & \multicolumn{3}{c|}{Sintel~\cite{butler2012naturalistic}}                                      & \multicolumn{3}{c}{TUM-dynamics~\cite{sturm2012benchmark}}                                                      \\ \cline{2-7} 
Method                   & ATE $\downarrow$ & RPE trans $\downarrow$ & RPE rot $\downarrow$ & ATE $\downarrow$ & RPE trans $\downarrow$ & \multicolumn{1}{c}{RPE rot $\downarrow$} \\ \hline
DROID-SLAM$^*$~\cite{teed2021droid}           & 0.175            & 0.084                  & 1.912                & -                & -                      & \multicolumn{1}{c}{-}                    \\
DPVO$^*$~\cite{teed2024deep}                 & 0.115            & 0.072                  & 1.975                & -                & -                      & \multicolumn{1}{c}{-}                    \\
ParticleSfM~\cite{zhao2022particlesfm}              & 0.129            & \textbf{0.031}         & {\ul 0.535}          & -                & -                      & \multicolumn{1}{c}{-}                    \\
LEAP-VO$^*$~\cite{chen2024leap}              & \textbf{0.089}   & 0.066                  & 1.250                & 0.068            & \textbf{0.008}         & \multicolumn{1}{c}{1.686}                \\ \hline
Robust-CVD~\cite{kopf2021rcvd}               & 0.360            & 0.154                  & 3.443                & 0.153            & 0.026                  & \multicolumn{1}{c}{3.528}                \\
CasualSAM~\cite{zhang2022structure}                & 0.141            & 0.035                  & 0.615                & 0.071            & 0.010                  & \multicolumn{1}{c}{1.712}                \\
DUSt3R~\cite{wang2024dust3r} w/ mask & 0.417            & 0.250                  & 5.796                & 0.083            & 0.017                  & \multicolumn{1}{c}{3.567}                \\
MonST3R~\cite{zhang2024monst3r}         & 0.108            & 0.042                  & 0.732                & {\ul 0.063}      & {\ul 0.009}            & \multicolumn{1}{c}{{\ul 1.217}}          \\ \hline
\textbf{Ours}                     & {\ul 0.090}      & {\ul 0.034}            & \textbf{0.312}       & \textbf{0.031}   & 0.011                  & \multicolumn{1}{c}{\textbf{0.426}}
\\ \hline

\end{tabular}
}
\label{tab:camera}
\end{table}
\begin{figure}[t]
    \centering
    \includegraphics[width=1.0\columnwidth]{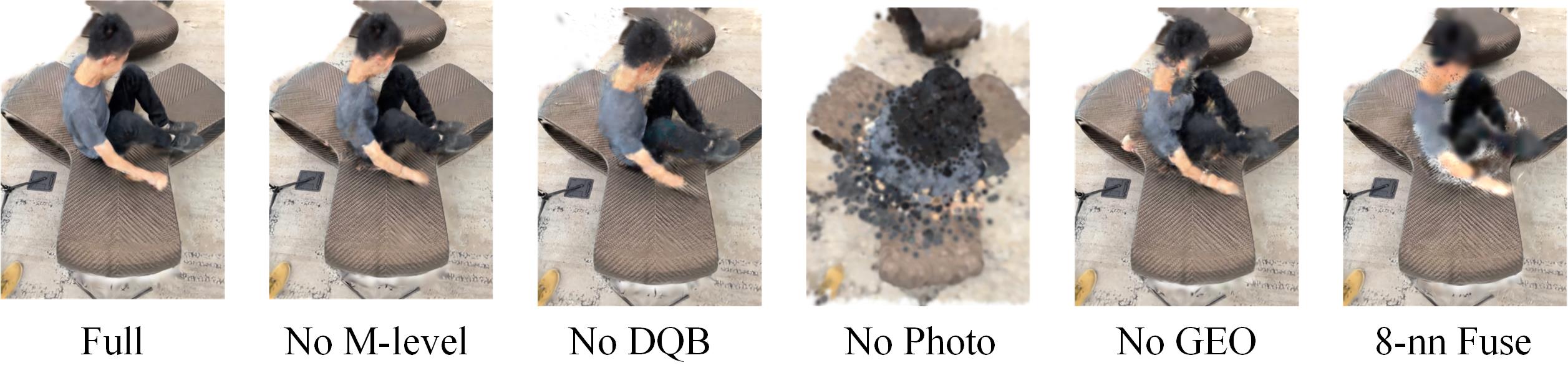}
    \caption{\small
    Visual comparison of ablation.
    }
    \label{fig:ablation}
    \vspace{-2em}
\end{figure}

\paragraph{Correspondence}
\begin{table}[b]
\vspace{-2em}
\caption{Correspondence on DyCheck~\cite{dycheck} with PCK-T @0.05\%}
\scalebox{0.7}{
\begin{tabular}{l|cccc}
\hline
Methods                           & \small{Nerfies}\cite{park2021nerfies}   & \small{HyperNeRf}\cite{park2021hypernerf}        & \small{Dyn. Gauss.}~\cite{luiten2023dynamic} & \small{4D Gauss.}~\cite{wu20234d} \\
\textit{PCK-T} $\uparrow$ & 0.4       & 0.453            & 0.079       & 0.073     \\ \hline
Methods                           & \small{CoTracker}\cite{cotracker} & \small{Gauss.Marbles}\cite{stearns2024dgmarbles} & \small{BootsTAPIR}~\cite{doersch2024bootstap}  & \textbf{Ours}      \\
\textit{PCK-T} $\uparrow$ & 0.803     & 0.806            & 0.779       & \textbf{0.824}     \\ \hline
\end{tabular}
\label{tab:pck}
}
\end{table}

One feature of \Ours is its ability to perform global fusion and provide dense correspondence. We quantitatively evaluate the correspondence tracking accuracy following DyCheck~\cite{dycheck} and Gaussian Marbles~\cite{stearns2024dgmarbles}. Tab.~\ref{tab:pck} shows our state-of-the-art accuracy. Notably, \Ours is optimized starting from BootsTAPIR~\cite{doersch2024bootstap} on DyCheck, and we observe a significant improvement over the raw tracker after reconstruction optimization.

\subsection{Ablation Study}
\label{sec:exp_abl}
\begin{table}[t]
\caption{Ablation study on different components of the system.}
\scalebox{0.8}{
\begin{tabular}{l|rrr}
\hline
{\color[HTML]{181A1B} Components} & \multicolumn{1}{l}{{\color[HTML]{181A1B} mPSNR}} & \multicolumn{1}{l}{{\color[HTML]{181A1B} mSSIM}} & \multicolumn{1}{l}{{\color[HTML]{181A1B} mLPIPs}} \\ \hline
\textbf{Full model}                        & \textbf{19.32}                                   & {\ul 0.706}                                      & \textbf{0.264}                                    \\
No node control                   & {\ul 19.28}                                      & \textbf{0.707}                                   & {\ul 0.267}                                       \\
No learnable skinning correction  & 19.27                                            & \textbf{0.707}                                   & {\ul 0.267}                                       \\
No dual quaternion blending       & 19.18                                            & 0.701                                            & 0.276                                             \\
No multi-level topology           & 19.14                                            & 0.701                                            & 0.270                                             \\
No geometric optimizaiton stage   & 18.85                                            & 0.693                                            & 0.287                                             \\
No photometric optimization stage & 13.71                                            & 0.480                                            & 0.763                                             \\
Only fuse 4 neighboring frames    & 16.96                                            & 0.663                                            & 0.344                                             \\
Only fuse 8 neighboring frames    & 17.26                                            & 0.664                                            & 0.346                                             \\ \hline
\end{tabular}
}
\label{tab:abl_comp}
\end{table}

\begin{table}[t]
\caption{Ablation study on different priors on DyCheck~\cite{dycheck}.}
\scalebox{0.7}{
\centering
\begin{tabular}{c|cccccc}
\hline
\multicolumn{1}{c|}{Tracker}          & \multicolumn{2}{c}{{\color[HTML]{181A1B} \textbf{BootsTAPIR}~\cite{doersch2024bootstap}}} & \multicolumn{2}{c}{{\color[HTML]{181A1B} CoTracker-v3~\cite{karaev2024cotracker3}}}      & \multicolumn{2}{c}{{\color[HTML]{181A1B} SpaTracker~\cite{SpatialTracker}}}        \\
\multicolumn{1}{c|}{Depth}            & {\color[HTML]{181A1B} mPSNR}  & {\color[HTML]{181A1B} mLPIPs}  & {\color[HTML]{181A1B} mPSNR} & {\color[HTML]{181A1B} mLPIPs} & {\color[HTML]{181A1B} mPSNR} & {\color[HTML]{181A1B} mLPIPs} \\ \hline
{\color[HTML]{181A1B} \textbf{LIDAR}} & 19.32                         & 0.264                          & 19.55                        & 0.243                         & 19.32                        & 0.259                         \\
{\color[HTML]{181A1B} Metric3D-v2~\cite{metric3dv2}}    & 17.05                         & 0.331                          & 17.02                        & 0.320                         & 17.60                        & 0.301                         \\
{\color[HTML]{181A1B} UniDepth~\cite{unidepth}}       & 17.12                         & 0.323                          & 17.42                        & 0.299                         & 17.61                        & 0.300                         \\ \hline
\end{tabular}

\label{tab:prior}
}
\end{table}

We assess the effects of different components in our system in Tab.~\ref{tab:abl_comp} and Fig.~\ref{fig:ablation}. We observe that both the geometric optimization and photometric optimization phases are critical. DQB contributes to smooth results, the multi-level topology pyramid enhances global rigidity and shape, and node control along with learnable skinning further improves the expressiveness of our system. Additionally, our system benefits from the global fusion of observations from every frame.
We also verify the effectiveness of the tracking loss $\mathcal{L}{\text{track}}$. When $\mc L_{\text{track}}$ is not used, the PCK-T accuracy decreases from $0.824$ to $0.737$.
In Tab.~\ref{tab:prior}, we study how different foundation models affect performance. Note that Metric3D-v2~\cite{metric3dv2} and UniDepth~\cite{unidepth} are entirely RGB-based and do not use LiDAR sensor information, leading to a reasonable decrease in performance.
We report more specifications of our system in Tab.~\ref{tab:others}, where we observe near real-time inference FPS and the compactness of the \Ours nodes compared to the actual foreground GS used to model the scene.
\begin{table}[]
\vspace{-1em}
\caption{\small More specs of \Ours on DyCheck~\cite{dycheck} (averaged)}
\captionsetup{skip=0pt}
\scalebox{0.8}{
\begin{tabular}{|c|c|c|c|}
\hline
FPS (2x res)                 & Num of fg GS               & Num of nodes                 & Ratio: \#GS/\#nodes             \\ \hline
\multicolumn{1}{|c|}{37.823} & \multicolumn{1}{c|}{106596} & \multicolumn{1}{c|}{3177} & \multicolumn{1}{c|}{46.105} \\ \hline
\end{tabular}
}
\label{tab:others}
\vspace{-1em}
\end{table}

\subsection{Applications}
\label{sec:exp_application}
In-the-wild 4D reconstruction enables many interesting applications, as shown in Fig.~\ref{fig:application}. For example, we can remove the moving foreground (Figure~\ref{fig:application}-A), or remove occluders in an extremely challenging cup-game video to look through and see where the ball goes (Figure~\ref{fig:application}-B). Video object segmentation from DEVA~\cite{deva} can be lifted and baked into 4D to produce novel view semantic videos (Figure~\ref{fig:application}-C). Finally, the 4D video can be edited in flexible ways, as shown in Figure~\ref{fig:application}-D. We believe that \Ours will provide the community with many more possibilities for future applications.

\vspace{-0.5em}
\section{Limitations and Conclusion}
\vspace{-0.5em}
\paragraph{Limitations}
While \Ours achieves state-of-the-art performance on standard benchmarks and can operate on some in-the-wild videos, several limitations remain. 
(1) Our method relies on accurate 2D long-term tracks and depth estimation, indicating that improvements in these areas are crucial for enhancing our performance. 
(2) Our current framework only reconstructs areas that are visible at some point in the video; it would be advantageous to incorporate large-scale 2D/video diffusion priors to hallucinate areas that are never visible. 
(3) Another important issue for future work is the correct modeling of lighting effects such as shadows, reflections, liquids, and changes in exposure. These effects cannot be explained by deformation alone and may cause artifacts in the background.

In summary, this paper takes a step toward reconstruction and rendering from monocular in-the-wild casual videos
We hope this small step could inspire future exploration toward understanding our dynamic physical world.

\paragraph{Acknowledgements}
The authors appreciate the support of the gift from AWS AI to Penn Engineering's ASSET Center for Trustworthy AI; 
and the support of  the following grants: 
NSF IIS-RI 2212433, NSF FRR 2220868 awarded to UPenn, 
ARL grant W911NF-21-2-0104 and a Vannevar Bush Faculty Fellowship awarded to Stanford University.

The authors thank Minh-Quan Viet Bui and the authors of DyBluRF, Xiaoming Zhao and the authors of PGDVS for providing their per-scene evaluation metrics on DyCheck dataset.

{\small
\bibliographystyle{ieeenat_fullname}
\bibliography{main.bib}

\begin{thebibliography}{121}
\providecommand{\natexlab}[1]{#1}
\providecommand{\url}[1]{\texttt{#1}}
\expandafter\ifx\csname urlstyle\endcsname\relax
  \providecommand{\doi}[1]{doi: #1}\else
  \providecommand{\doi}{doi: \begingroup \urlstyle{rm}\Url}\fi

\bibitem[Athar et~al.(2022)Athar, Xu, Sunkavalli, Shechtman, and Shu]{athar2022rignerf}
ShahRukh Athar, Zexiang Xu, Kalyan Sunkavalli, Eli Shechtman, and Zhixin Shu.
\newblock Rignerf: Fully controllable neural 3d portraits.
\newblock In \emph{Proceedings of the IEEE/CVF Conference on Computer Vision and Pattern Recognition}, pages 20364--20373, 2022.

\bibitem[Attal et~al.(2023)Attal, Huang, Richardt, Zollhoefer, Kopf, O'Toole, and Kim]{attal2023hyperreel}
Benjamin Attal, Jia-Bin Huang, Christian Richardt, Michael Zollhoefer, Johannes Kopf, Matthew O'Toole, and Changil Kim.
\newblock Hyperreel: High-fidelity 6-dof video with ray-conditioned sampling.
\newblock In \emph{Proceedings of the IEEE/CVF Conference on Computer Vision and Pattern Recognition (CVPR)}, 2023.

\bibitem[Bansal et~al.(2020)Bansal, Vo, Sheikh, Ramanan, and Narasimhan]{bansal20204d}
Aayush Bansal, Minh Vo, Yaser Sheikh, Deva Ramanan, and Srinivasa Narasimhan.
\newblock 4d visualization of dynamic events from unconstrained multi-view videos.
\newblock In \emph{Proceedings of the IEEE/CVF Conference on Computer Vision and Pattern Recognition (CVPR)}, 2020.

\bibitem[Barron et~al.(2021)Barron, Mildenhall, Tancik, Hedman, Martin-Brualla, and Srinivasan]{barron2021mip}
Jonathan~T Barron, Ben Mildenhall, Matthew Tancik, Peter Hedman, Ricardo Martin-Brualla, and Pratul~P Srinivasan.
\newblock Mip-nerf: A multiscale representation for anti-aliasing neural radiance fields.
\newblock In \emph{Proceedings of the IEEE/CVF International Conference on Computer Vision}, pages 5855--5864, 2021.

\bibitem[Bemana et~al.(2020)Bemana, Myszkowski, Seidel, and Ritschel]{bemana2020xfields}
Mojtaba Bemana, Karol Myszkowski, Hans-Peter Seidel, and Tobias Ritschel.
\newblock X-fields: Implicit neural view-, light- and time-image interpolation.
\newblock \emph{SIGGRAPH Asia}, 2020.

\bibitem[Bhat et~al.(2023)Bhat, Birkl, Wofk, Wonka, and M{\"u}ller]{zoedepth}
Shariq~Farooq Bhat, Reiner Birkl, Diana Wofk, Peter Wonka, and Matthias M{\"u}ller.
\newblock Zoedepth: Zero-shot transfer by combining relative and metric depth.
\newblock \emph{arXiv preprint arXiv:2302.12288}, 2023.

\bibitem[Blanz and Vetter(1999)]{blanz1999morphable}
Volker Blanz and Thomas Vetter.
\newblock A morphable model for the synthesis of 3d faces.
\newblock In \emph{Proceedings of the 26th Annual Conference on Computer Graphics and Interactive Techniques}, pages 187--194, 1999.

\bibitem[Bogo et~al.(2016)Bogo, Kanazawa, Lassner, Gehler, Romero, and Black]{bogo2016keep}
Federica Bogo, Angjoo Kanazawa, Christoph Lassner, Peter Gehler, Javier Romero, and Michael~J Black.
\newblock Keep it smpl: Automatic estimation of 3d human pose and shape from a single image.
\newblock In \emph{European Conference on Computer Vision}, pages 561--578. Springer, 2016.

\bibitem[Bommasani et~al.(2021)Bommasani, Hudson, Adeli, Altman, Arora, von Arx, Bernstein, Bohg, Bosselut, Brunskill, et~al.]{bommasani2021opportunities}
Rishi Bommasani, Drew~A Hudson, Ehsan Adeli, Russ Altman, Simran Arora, Sydney von Arx, Michael~S Bernstein, Jeannette Bohg, Antoine Bosselut, Emma Brunskill, et~al.
\newblock On the opportunities and risks of foundation models.
\newblock \emph{arXiv preprint arXiv:2108.07258}, 2021.

\bibitem[Bozic et~al.(2020)Bozic, Palafox, Zoll{\"o}fer, Dai, Thies, and Nie{\ss}ner]{bozic2020neural}
Aljaz Bozic, Pablo Palafox, Michael Zoll{\"o}fer, Angela Dai, Justus Thies, and Matthias Nie{\ss}ner.
\newblock Neural non-rigid tracking.
\newblock In \emph{Advances in Neural Information Processing Systems}, pages 18765--18775, 2020.

\bibitem[Bui et~al.(2023)Bui, Park, Oh, and Kim]{bui2023dyblurf}
Minh-Quan~Viet Bui, Jongmin Park, Jihyong Oh, and Munchurl Kim.
\newblock Dyblurf: Dynamic deblurring neural radiance fields for blurry monocular video.
\newblock \emph{arXiv preprint arXiv:2312.13528}, 2023.

\bibitem[Butler et~al.(2012)Butler, Wulff, Stanley, and Black]{butler2012naturalistic}
Daniel~J Butler, Jonas Wulff, Garrett~B Stanley, and Michael~J Black.
\newblock A naturalistic open source movie for optical flow evaluation.
\newblock In \emph{Computer Vision--ECCV 2012: 12th European Conference on Computer Vision, Florence, Italy, October 7-13, 2012, Proceedings, Part VI 12}, pages 611--625. Springer, 2012.

\bibitem[Cao and Johnson(2023)]{cao2023hexplane}
Ang Cao and Justin Johnson.
\newblock Hexplane: A fast representation for dynamic scenes.
\newblock \emph{arXiv preprint arXiv:2301.09632}, 2023.

\bibitem[Chen et~al.(2022)Chen, Xu, Geiger, Yu, and Su]{chen2022tensorf}
Anpei Chen, Zexiang Xu, Andreas Geiger, Jingyi Yu, and Hao Su.
\newblock Tensorf: Tensorial radiance fields.
\newblock In \emph{Computer Vision--ECCV 2022: 17th European Conference, Tel Aviv, Israel, October 23--27, 2022, Proceedings, Part XXXII}, pages 333--350. Springer, 2022.

\bibitem[Chen et~al.(2024)Chen, Chen, Wang, and Pollefeys]{chen2024leap}
Weirong Chen, Le Chen, Rui Wang, and Marc Pollefeys.
\newblock Leap-vo: Long-term effective any point tracking for visual odometry.
\newblock In \emph{IEEE/CVF Conference on Computer Vision and Pattern Recognition (CVPR)}, 2024.

\bibitem[Chen et~al.(2023)Chen, Wang, Li, Xiao, Zhang, Yao, and Liu]{chen2023monogaussianavatar}
Yufan Chen, Lizhen Wang, Qijing Li, Hongjiang Xiao, Shengping Zhang, Hongxun Yao, and Yebin Liu.
\newblock Monogaussianavatar: Monocular gaussian point-based head avatar.
\newblock \emph{arXiv preprint arXiv:2312.04558}, 2023.

\bibitem[Cheng et~al.(2023)Cheng, Oh, Price, Schwing, and Lee]{deva}
Ho~Kei Cheng, Seoung~Wug Oh, Brian Price, Alexander Schwing, and Joon-Young Lee.
\newblock Tracking anything with decoupled video segmentation.
\newblock In \emph{ICCV}, 2023.

\bibitem[Chu et~al.(2024)Chu, Ke, and Fragkiadaki]{chu2024dreamscene4d}
Wen-Hsuan Chu, Lei Ke, and Katerina Fragkiadaki.
\newblock Dreamscene4d: Dynamic multi-object scene generation from monocular videos.
\newblock \emph{arXiv preprint arXiv:2405.02280}, 2024.

\bibitem[Curless and Levoy(1996)]{curless1996volumetric}
Brian Curless and Marc Levoy.
\newblock A volumetric method for building complex models from range images.
\newblock In \emph{Proceedings of the 23rd annual conference on Computer graphics and interactive techniques}, pages 303--312, 1996.

\bibitem[Daniilidis(1999)]{daniilidis1999hand}
Konstantinos Daniilidis.
\newblock Hand-eye calibration using dual quaternions.
\newblock \emph{The International Journal of Robotics Research}, 18\penalty0 (3):\penalty0 286--298, 1999.

\bibitem[Das et~al.(2023)Das, Wewer, Yunus, Ilg, and Lenssen]{das2023neural}
Devikalyan Das, Christopher Wewer, Raza Yunus, Eddy Ilg, and Jan~Eric Lenssen.
\newblock Neural parametric gaussians for monocular non-rigid object reconstruction.
\newblock \emph{arXiv preprint arXiv:2312.01196}, 2023.

\bibitem[Doersch et~al.(2024)Doersch, Luc, Yang, Gokay, Koppula, Gupta, Heyward, Rocco, Goroshin, Carreira, and Zisserman]{doersch2024bootstap}
Carl Doersch, Pauline Luc, Yi Yang, Dilara Gokay, Skanda Koppula, Ankush Gupta, Joseph Heyward, Ignacio Rocco, Ross Goroshin, Joao Carreira, and Andrew Zisserman.
\newblock Bootstap: Bootstrapped training for tracking-any-point.
\newblock \emph{Asian Conference on Computer Vision}, 2024.

\bibitem[Dou et~al.(2015)Dou, Taylor, Fuchs, Fitzgibbon, and Izadi]{dou20153d}
Mingsong Dou, Jonathan Taylor, Henry Fuchs, Andrew Fitzgibbon, and Shahram Izadi.
\newblock 3d scanning deformable objects with a single rgbd sensor.
\newblock In \emph{Proceedings of the IEEE Conference on Computer Vision and Pattern Recognition}, pages 493--501, 2015.

\bibitem[Du et~al.(2021)Du, Zhang, Yu, Tenenbaum, and Wu]{du2021neural}
Yilun Du, Yinan Zhang, Hong-Xing Yu, Joshua~B Tenenbaum, and Jiajun Wu.
\newblock Neural radiance flow for 4d view synthesis and video processing.
\newblock In \emph{2021 IEEE/CVF International Conference on Computer Vision (ICCV)}, pages 14304--14314. IEEE Computer Society, 2021.

\bibitem[Duan et~al.(2024)Duan, Wei, Dai, He, Chen, and Chen]{duan20244d}
Yuanxing Duan, Fangyin Wei, Qiyu Dai, Yuhang He, Wenzheng Chen, and Baoquan Chen.
\newblock 4d gaussian splatting: Towards efficient novel view synthesis for dynamic scenes.
\newblock \emph{arXiv preprint arXiv:2402.03307}, 2024.

\bibitem[Duisterhof et~al.(2023)Duisterhof, Mandi, Yao, Liu, Shou, Song, and Ichnowski]{duisterhof2023md}
Bardienus~P Duisterhof, Zhao Mandi, Yunchao Yao, Jia-Wei Liu, Mike~Zheng Shou, Shuran Song, and Jeffrey Ichnowski.
\newblock Md-splatting: Learning metric deformation from 4d gaussians in highly deformable scenes.
\newblock \emph{arXiv preprint arXiv:2312.00583}, 2023.

\bibitem[Fang et~al.(2022)Fang, Yi, Wang, Xie, Zhang, Liu, Nie{\ss}ner, and Tian]{fang2022fast}
Jiemin Fang, Taoran Yi, Xinggang Wang, Lingxi Xie, Xiaopeng Zhang, Wenyu Liu, Matthias Nie{\ss}ner, and Qi Tian.
\newblock Fast dynamic radiance fields with time-aware neural voxels.
\newblock In \emph{SIGGRAPH Asia 2022 Conference Papers}, pages 1--9, 2022.

\bibitem[Fridovich-Keil et~al.(2023)Fridovich-Keil, Meanti, Warburg, Recht, and Kanazawa]{fridovich2023k}
Sara Fridovich-Keil, Giacomo Meanti, Frederik Warburg, Benjamin Recht, and Angjoo Kanazawa.
\newblock K-planes: Explicit radiance fields in space, time, and appearance.
\newblock \emph{arXiv preprint arXiv:2301.10241}, 2023.

\bibitem[Gao et~al.(2021)Gao, Saraf, Kopf, and Huang]{gao2021dynamic}
Chen Gao, Ayush Saraf, Johannes Kopf, and Jia-Bin Huang.
\newblock Dynamic view synthesis from dynamic monocular video.
\newblock In \emph{Proceedings of the IEEE/CVF International Conference on Computer Vision}, pages 5712--5721, 2021.

\bibitem[Gao et~al.(2022)Gao, Li, Tulsiani, Russell, and Kanazawa]{dycheck}
Hang Gao, Ruilong Li, Shubham Tulsiani, Bryan Russell, and Angjoo Kanazawa.
\newblock Monocular dynamic view synthesis: A reality check.
\newblock \emph{Advances in Neural Information Processing Systems}, 35:\penalty0 33768--33780, 2022.

\bibitem[Gao and Tedrake(2018)]{gao2018surfelwarp}
Wei Gao and Russ Tedrake.
\newblock Surfelwarp: Efficient non-volumetric single view dynamic reconstruction.
\newblock In \emph{Robotics: Science and Systems (RSS)}, 2018.

\bibitem[Harley et~al.(2022)Harley, Fang, and Fragkiadaki]{pips}
Adam~W. Harley, Zhaoyuan Fang, and Katerina Fragkiadaki.
\newblock Particle video revisited: {T}racking through occlusions using point trajectories.
\newblock 2022.

\bibitem[Hu et~al.(2023)Hu, Zhang, Zhang, Zhou, Liu, Zhang, and Nie]{hu2023gaussianavatar}
Liangxiao Hu, Hongwen Zhang, Yuxiang Zhang, Boyao Zhou, Boning Liu, Shengping Zhang, and Liqiang Nie.
\newblock Gaussianavatar: Towards realistic human avatar modeling from a single video via animatable 3d gaussians.
\newblock \emph{arXiv preprint arXiv:2312.02134}, 2023.

\bibitem[Hu et~al.(2024{\natexlab{a}})Hu, Yin, Zhang, Cai, Long, Chen, Wang, Yu, Shen, and Shen]{metric3dv2}
Mu Hu, Wei Yin, Chi Zhang, Zhipeng Cai, Xiaoxiao Long, Hao Chen, Kaixuan Wang, Gang Yu, Chunhua Shen, and Shaojie Shen.
\newblock Metric3d v2: A versatile monocular geometric foundation model for zero-shot metric depth and surface normal estimation.
\newblock 2024{\natexlab{a}}.

\bibitem[Hu and Liu(2023)]{hu2023gauhuman}
Shoukang Hu and Ziwei Liu.
\newblock Gauhuman: Articulated gaussian splatting from monocular human videos.
\newblock \emph{arXiv preprint arXiv:2312.02973}, 2023.

\bibitem[Hu et~al.(2024{\natexlab{b}})Hu, Gao, Li, Zhao, Cun, Zhang, Quan, and Shan]{depthcrafter}
Wenbo Hu, Xiangjun Gao, Xiaoyu Li, Sijie Zhao, Xiaodong Cun, Yong Zhang, Long Quan, and Ying Shan.
\newblock Depthcrafter: Generating consistent long depth sequences for open-world videos.
\newblock \emph{arXiv preprint arXiv:2409.02095}, 2024{\natexlab{b}}.

\bibitem[Huang et~al.(2023)Huang, Sun, Yang, Lyu, Cao, and Qi]{huang2023sc}
Yi-Hua Huang, Yang-Tian Sun, Ziyi Yang, Xiaoyang Lyu, Yan-Pei Cao, and Xiaojuan Qi.
\newblock Sc-gs: Sparse-controlled gaussian splatting for editable dynamic scenes.
\newblock \emph{arXiv preprint arXiv:2312.14937}, 2023.

\bibitem[Jia()]{jia2013dual}
Yan-Bin Jia.
\newblock Dual quaternions.

\bibitem[Kappel et~al.(2024)Kappel, Hahlbohm, Scholz, Castillo, Theobalt, Eisemann, Golyanik, and Magnor]{kappel2024d}
Moritz Kappel, Florian Hahlbohm, Timon Scholz, Susana Castillo, Christian Theobalt, Martin Eisemann, Vladislav Golyanik, and Marcus Magnor.
\newblock D-npc: Dynamic neural point clouds for non-rigid view synthesis from monocular video.
\newblock \emph{arXiv preprint arXiv:2406.10078}, 2024.

\bibitem[Karaev et~al.(2023)Karaev, Rocco, Graham, Neverova, Vedaldi, and Rupprecht]{cotracker}
Nikita Karaev, Ignacio Rocco, Benjamin Graham, Natalia Neverova, Andrea Vedaldi, and Christian Rupprecht.
\newblock Cotracker: It is better to track together.
\newblock \emph{arXiv preprint arXiv:2307.07635}, 2023.

\bibitem[Karaev et~al.(2024)Karaev, Makarov, Wang, Neverova, Vedaldi, and Rupprecht]{karaev2024cotracker3}
Nikita Karaev, Iurii Makarov, Jianyuan Wang, Natalia Neverova, Andrea Vedaldi, and Christian Rupprecht.
\newblock Cotracker3: Simpler and better point tracking by pseudo-labelling real videos.
\newblock 2024.

\bibitem[Katsumata et~al.(2023)Katsumata, Vo, and Nakayama]{katsumata2023efficient}
Kai Katsumata, Duc~Minh Vo, and Hideki Nakayama.
\newblock An efficient 3d gaussian representation for monocular/multi-view dynamic scenes.
\newblock \emph{arXiv preprint arXiv:2311.12897}, 2023.

\bibitem[Kavan et~al.(2007)Kavan, Collins, {\v{Z}}{\'a}ra, and O'Sullivan]{DQB}
Ladislav Kavan, Steven Collins, Ji{\v{r}}{\'\i} {\v{Z}}{\'a}ra, and Carol O'Sullivan.
\newblock Skinning with dual quaternions.
\newblock In \emph{Proceedings of the 2007 symposium on Interactive 3D graphics and games}, pages 39--46, 2007.

\bibitem[Kerbl et~al.(2023)Kerbl, Kopanas, Leimk{\"u}hler, and Drettakis]{kerbl20233d}
Bernhard Kerbl, Georgios Kopanas, Thomas Leimk{\"u}hler, and George Drettakis.
\newblock 3d gaussian splatting for real-time radiance field rendering.
\newblock 2023.

\bibitem[Keselman and Hebert(2022)]{keselman2022approximate}
Leonid Keselman and Martial Hebert.
\newblock Approximate differentiable rendering with algebraic surfaces.
\newblock In \emph{European Conference on Computer Vision}, pages 596--614. Springer, 2022.

\bibitem[Keselman and Hebert(2023)]{keselman2023flexible}
Leonid Keselman and Martial Hebert.
\newblock Flexible techniques for differentiable rendering with 3d gaussians.
\newblock \emph{arXiv preprint arXiv:2308.14737}, 2023.

\bibitem[Kirillov et~al.(2023)Kirillov, Mintun, Ravi, Mao, Rolland, Gustafson, Xiao, Whitehead, Berg, Lo, et~al.]{kirillov2023segment}
Alexander Kirillov, Eric Mintun, Nikhila Ravi, Hanzi Mao, Chloe Rolland, Laura Gustafson, Tete Xiao, Spencer Whitehead, Alexander~C Berg, Wan-Yen Lo, et~al.
\newblock Segment anything.
\newblock In \emph{Proceedings of the IEEE/CVF International Conference on Computer Vision}, pages 4015--4026, 2023.

\bibitem[Kocabas et~al.(2023)Kocabas, Chang, Gabriel, Tuzel, and Ranjan]{kocabas2023hugs}
Muhammed Kocabas, Jen-Hao~Rick Chang, James Gabriel, Oncel Tuzel, and Anurag Ranjan.
\newblock Hugs: Human gaussian splats.
\newblock \emph{arXiv preprint arXiv:2311.17910}, 2023.

\bibitem[Kopf et~al.(2021)Kopf, Rong, and Huang]{kopf2021rcvd}
Johannes Kopf, Xuejian Rong, and Jia-Bin Huang.
\newblock Robust consistent video depth estimation.
\newblock In \emph{IEEE/CVF Conference on Computer Vision and Pattern Recognition}, 2021.

\bibitem[Kratimenos et~al.(2023)Kratimenos, Lei, and Daniilidis]{kratimenos2023dynmf}
Agelos Kratimenos, Jiahui Lei, and Kostas Daniilidis.
\newblock Dynmf: Neural motion factorization for real-time dynamic view synthesis with 3d gaussian splatting.
\newblock \emph{arXiv preprint arXiv:2312.00112}, 2023.

\bibitem[Lee et~al.(2023)Lee, Zhang, Blackburn-Matzen, Niklaus, Zhang, Huang, and Liu]{lee2023fast}
Yao-Chih Lee, Zhoutong Zhang, Kevin Blackburn-Matzen, Simon Niklaus, Jianming Zhang, Jia-Bin Huang, and Feng Liu.
\newblock Fast view synthesis of casual videos.
\newblock \emph{arXiv preprint arXiv:2312.02135}, 2023.

\bibitem[Lei et~al.(2023)Lei, Wang, Pavlakos, Liu, and Daniilidis]{lei2023gart}
Jiahui Lei, Yufu Wang, Georgios Pavlakos, Lingjie Liu, and Kostas Daniilidis.
\newblock Gart: Gaussian articulated template models.
\newblock \emph{arXiv preprint arXiv:2311.16099}, 2023.

\bibitem[Li et~al.(2008)Li, Sumner, and Pauly]{li2008global}
Hao Li, Robert~W Sumner, and Mark Pauly.
\newblock Global correspondence optimization for non-rigid registration of depth scans.
\newblock \emph{Computer Graphics Forum}, 27\penalty0 (5):\penalty0 1421--1430, 2008.

\bibitem[Li et~al.(2023{\natexlab{a}})Li, Tao, Yang, and Yang]{li2023human101}
Mingwei Li, Jiachen Tao, Zongxin Yang, and Yi Yang.
\newblock Human101: Training 100+ fps human gaussians in 100s from 1 view.
\newblock \emph{arXiv preprint arXiv:2312.15258}, 2023{\natexlab{a}}.

\bibitem[Li et~al.(2022)Li, Slavcheva, Zollhoefer, Green, Lassner, Kim, Schmidt, Lovegrove, Goesele, Newcombe, et~al.]{li2022neural3d}
Tianye Li, Mira Slavcheva, Michael Zollhoefer, Simon Green, Christoph Lassner, Changil Kim, Tanner Schmidt, Steven Lovegrove, Michael Goesele, Richard Newcombe, et~al.
\newblock Neural 3d video synthesis from multi-view video.
\newblock In \emph{Proceedings of the IEEE/CVF Conference on Computer Vision and Pattern Recognition (CVPR)}, 2022.

\bibitem[Li et~al.(2021)Li, Niklaus, Snavely, and Wang]{li2021neural}
Zhengqi Li, Simon Niklaus, Noah Snavely, and Oliver Wang.
\newblock Neural scene flow fields for space-time view synthesis of dynamic scenes.
\newblock In \emph{Proceedings of the IEEE/CVF Conference on Computer Vision and Pattern Recognition}, pages 6498--6508, 2021.

\bibitem[Li et~al.(2023{\natexlab{b}})Li, Chen, Li, and Xu]{li2023spacetime}
Zhan Li, Zhang Chen, Zhong Li, and Yi Xu.
\newblock Spacetime gaussian feature splatting for real-time dynamic view synthesis.
\newblock \emph{arXiv preprint arXiv:2312.16812}, 2023{\natexlab{b}}.

\bibitem[Li et~al.(2023{\natexlab{c}})Li, Wang, Cole, Tucker, and Snavely]{li2023dynibar}
Zhengqi Li, Qianqian Wang, Forrester Cole, Richard Tucker, and Noah Snavely.
\newblock Dynibar: Neural dynamic image-based rendering, 2023{\natexlab{c}}.

\bibitem[Liang et~al.(2023)Liang, Khan, Li, Nguyen-Phuoc, Lanman, Tompkin, and Xiao]{liang2023gaufre}
Yiqing Liang, Numair Khan, Zhengqin Li, Thu Nguyen-Phuoc, Douglas Lanman, James Tompkin, and Lei Xiao.
\newblock Gaufre: Gaussian deformation fields for real-time dynamic novel view synthesis.
\newblock \emph{arXiv preprint arXiv:2312.11458}, 2023.

\bibitem[Lin et~al.(2023{\natexlab{a}})Lin, Peng, Xu, Xie, He, Bao, and Zhou]{lin2023im4d}
Haotong Lin, Sida Peng, Zhen Xu, Tao Xie, Xingyi He, Hujun Bao, and Xiaowei Zhou.
\newblock High-fidelity and real-time novel view synthesis for dynamic scenes.
\newblock In \emph{SIGGRAPH Asia Conference Proceedings}, 2023{\natexlab{a}}.

\bibitem[Lin et~al.(2023{\natexlab{b}})Lin, Dai, Zhu, and Yao]{lin2023gaussian}
Youtian Lin, Zuozhuo Dai, Siyu Zhu, and Yao Yao.
\newblock Gaussian-flow: 4d reconstruction with dynamic 3d gaussian particle.
\newblock \emph{arXiv preprint arXiv:2312.03431}, 2023{\natexlab{b}}.

\bibitem[Liu et~al.(2023{\natexlab{a}})Liu, Li, Li, and Lee]{liu2023improved}
Haotian Liu, Chunyuan Li, Yuheng Li, and Yong~Jae Lee.
\newblock Improved baselines with visual instruction tuning.
\newblock \emph{arXiv preprint arXiv:2310.03744}, 2023{\natexlab{a}}.

\bibitem[Liu et~al.(2024{\natexlab{a}})Liu, Li, Wu, and Lee]{liu2024visual}
Haotian Liu, Chunyuan Li, Qingyang Wu, and Yong~Jae Lee.
\newblock Visual instruction tuning.
\newblock \emph{Advances in neural information processing systems}, 36, 2024{\natexlab{a}}.

\bibitem[Liu et~al.(2024{\natexlab{b}})Liu, Liu, Wang, Lv, Wang, Wang, and Hou]{liu2024modgs}
Qingming Liu, Yuan Liu, Jiepeng Wang, Xianqiang Lv, Peng Wang, Wenping Wang, and Junhui Hou.
\newblock {MoDGS: Dynamic Gaussian Splatting from Causally-Captured Monocular Videos}.
\newblock \emph{arXiv preprint arXiv:2406.00434}, 2024{\natexlab{b}}.

\bibitem[Liu et~al.(2024{\natexlab{c}})Liu, Wu, Liu, Liu, Wu, Zhao, Feng, Ding, and Wang]{liu2024gva}
Xinqi Liu, Chenming Wu, Jialun Liu, Xing Liu, Jinbo Wu, Chen Zhao, Haocheng Feng, Errui Ding, and Jingdong Wang.
\newblock Gva: Reconstructing vivid 3d gaussian avatars from monocular videos, 2024{\natexlab{c}}.

\bibitem[Liu et~al.(2023{\natexlab{b}})Liu, Gao, Meuleman, Tseng, Saraf, Kim, Chuang, Kopf, and Huang]{rodynrf}
Yu-Lun Liu, Chen Gao, Andreas Meuleman, Hung-Yu Tseng, Ayush Saraf, Changil Kim, Yung-Yu Chuang, Johannes Kopf, and Jia-Bin Huang.
\newblock Robust dynamic radiance fields.
\newblock In \emph{Proceedings of the IEEE/CVF Conference on Computer Vision and Pattern Recognition}, pages 13--23, 2023{\natexlab{b}}.

\bibitem[Luiten et~al.(2023)Luiten, Kopanas, Leibe, and Ramanan]{luiten2023dynamic}
Jonathon Luiten, Georgios Kopanas, Bastian Leibe, and Deva Ramanan.
\newblock Dynamic 3d gaussians: Tracking by persistent dynamic view synthesis.
\newblock \emph{arXiv preprint arXiv:2308.09713}, 2023.

\bibitem[Miao et~al.(2024)Miao, Bai, Duan, Huang, Wan, Long, and Zheng]{miao2024ctnerf}
Xingyu Miao, Yang Bai, Haoran Duan, Yawen Huang, Fan Wan, Yang Long, and Yefeng Zheng.
\newblock Ctnerf: Cross-time transformer for dynamic neural radiance field from monocular video.
\newblock \emph{arXiv preprint arXiv:2401.04861}, 2024.

\bibitem[Mildenhall et~al.(2021)Mildenhall, Srinivasan, Tancik, Barron, Ramamoorthi, and Ng]{mildenhall2021nerf}
Ben Mildenhall, Pratul~P Srinivasan, Matthew Tancik, Jonathan~T Barron, Ravi Ramamoorthi, and Ren Ng.
\newblock Nerf: Representing scenes as neural radiance fields for view synthesis.
\newblock \emph{Communications of the ACM}, 65\penalty0 (1):\penalty0 99--106, 2021.

\bibitem[M{\"u}ller et~al.(2022)M{\"u}ller, Evans, Schied, and Keller]{muller2022instant}
Thomas M{\"u}ller, Alex Evans, Christoph Schied, and Alexander Keller.
\newblock Instant neural graphics primitives with a multiresolution hash encoding.
\newblock \emph{arXiv preprint arXiv:2201.05989}, 2022.

\bibitem[Newcombe et~al.(2015)Newcombe, Fox, and Seitz]{newcombe2015dynamicfusion}
Richard~A Newcombe, Dieter Fox, and Steven~M Seitz.
\newblock Dynamicfusion: Reconstruction and tracking of non-rigid scenes in real-time.
\newblock In \emph{Proceedings of the IEEE conference on computer vision and pattern recognition}, pages 343--352, 2015.

\bibitem[OpenAI(2023)]{openai2023gpt4}
OpenAI.
\newblock Gpt-4 technical report, 2023.
\newblock https://openai.com/research/gpt-4.

\bibitem[Oquab et~al.(2023)Oquab, Darcet, Moutakanni, Vo, Szafraniec, Khalidov, Fernandez, Haziza, Massa, El-Nouby, et~al.]{dinov2}
Maxime Oquab, Timoth{\'e}e Darcet, Th{\'e}o Moutakanni, Huy Vo, Marc Szafraniec, Vasil Khalidov, Pierre Fernandez, Daniel Haziza, Francisco Massa, Alaaeldin El-Nouby, et~al.
\newblock Dinov2: Learning robust visual features without supervision.
\newblock \emph{arXiv preprint arXiv:2304.07193}, 2023.

\bibitem[Park et~al.(2021{\natexlab{a}})Park, Sinha, Barron, Bouaziz, Goldman, Seitz, and Martin-Brualla]{park2021nerfies}
Keunhong Park, Utkarsh Sinha, Jonathan~T Barron, Sofien Bouaziz, Dan~B Goldman, Steven~M Seitz, and Ricardo Martin-Brualla.
\newblock Nerfies: Deformable neural radiance fields.
\newblock In \emph{Proceedings of the IEEE/CVF International Conference on Computer Vision}, pages 5865--5874, 2021{\natexlab{a}}.

\bibitem[Park et~al.(2021{\natexlab{b}})Park, Sinha, Hedman, Barron, Bouaziz, Goldman, Martin-Brualla, and Seitz]{park2021hypernerf}
Keunhong Park, Utkarsh Sinha, Peter Hedman, Jonathan~T Barron, Sofien Bouaziz, Dan~B Goldman, Ricardo Martin-Brualla, and Steven~M Seitz.
\newblock Hypernerf: A higher-dimensional representation for topologically varying neural radiance fields.
\newblock \emph{arXiv preprint arXiv:2106.13228}, 2021{\natexlab{b}}.

\bibitem[Piccinelli et~al.(2024)Piccinelli, Yang, Sakaridis, Segu, Li, Van~Gool, and Yu]{unidepth}
Luigi Piccinelli, Yung-Hsu Yang, Christos Sakaridis, Mattia Segu, Siyuan Li, Luc Van~Gool, and Fisher Yu.
\newblock Unidepth: Universal monocular metric depth estimation.
\newblock \emph{arXiv preprint arXiv:2403.18913}, 2024.

\bibitem[Pont-Tuset et~al.(2017)Pont-Tuset, Perazzi, Caelles, Arbel{\'a}ez, Sorkine-Hornung, and Van~Gool]{davis}
Jordi Pont-Tuset, Federico Perazzi, Sergi Caelles, Pablo Arbel{\'a}ez, Alex Sorkine-Hornung, and Luc Van~Gool.
\newblock The 2017 davis challenge on video object segmentation.
\newblock \emph{arXiv preprint arXiv:1704.00675}, 2017.

\bibitem[Pumarola et~al.(2021)Pumarola, Corona, Pons-Moll, and Moreno-Noguer]{pumarola2021d}
Albert Pumarola, Enric Corona, Gerard Pons-Moll, and Francesc Moreno-Noguer.
\newblock D-nerf: Neural radiance fields for dynamic scenes.
\newblock In \emph{Proceedings of the IEEE/CVF Conference on Computer Vision and Pattern Recognition}, pages 10318--10327, 2021.

\bibitem[Qian et~al.(2023)Qian, Wang, Mihajlovic, Geiger, and Tang]{qian20233dgs}
Zhiyin Qian, Shaofei Wang, Marko Mihajlovic, Andreas Geiger, and Siyu Tang.
\newblock 3dgs-avatar: Animatable avatars via deformable 3d gaussian splatting.
\newblock \emph{arXiv preprint arXiv:2312.09228}, 2023.

\bibitem[Radford et~al.(2021)Radford, Kim, Hallacy, Ramesh, Goh, Agarwal, Sastry, Askell, Mishkin, Clark, et~al.]{radford2021learning}
Alec Radford, Jong~Wook Kim, Chris Hallacy, Aditya Ramesh, Gabriel Goh, Sandhini Agarwal, Girish Sastry, Amanda Askell, Pamela Mishkin, Jack Clark, et~al.
\newblock Learning transferable visual models from natural language supervision.
\newblock In \emph{International conference on machine learning}, pages 8748--8763. PMLR, 2021.

\bibitem[Ramakrishna et~al.(2012)Ramakrishna, Kanade, and Sheikh]{ramakrishna2012reconstructing}
Varun Ramakrishna, Takeo Kanade, and Yaser Sheikh.
\newblock Reconstructing 3d human pose from 2d image landmarks.
\newblock In \emph{European Conference on Computer Vision}, pages 573--586. Springer, 2012.

\bibitem[Rivero et~al.(2024)Rivero, Athar, Shu, and Samaras]{rivero2024rig3dgs}
Alfredo Rivero, ShahRukh Athar, Zhixin Shu, and Dimitris Samaras.
\newblock Rig3dgs: Creating controllable portraits from casual monocular videos.
\newblock \emph{arXiv preprint arXiv:2402.03723}, 2024.

\bibitem[Seidenschwarz et~al.(2024)Seidenschwarz, Zhou, Duisterhof, Ramanan, and Leal-Taix{\'e}]{seidenschwarz2024dynomo}
Jenny Seidenschwarz, Qunjie Zhou, Bardienus Duisterhof, Deva Ramanan, and Laura Leal-Taix{\'e}.
\newblock Dynomo: Online point tracking by dynamic online monocular gaussian reconstruction.
\newblock \emph{arXiv preprint arXiv:2409.02104}, 2024.

\bibitem[Shao et~al.(2024)Shao, Wang, Li, Wang, Lin, Zhang, Fan, and Wang]{shao2024splattingavatar}
Zhijing Shao, Zhaolong Wang, Zhuang Li, Duotun Wang, Xiangru Lin, Yu Zhang, Mingming Fan, and Zeyu Wang.
\newblock Splattingavatar: Realistic real-time human avatars with mesh-embedded gaussian splatting.
\newblock \emph{arXiv preprint arXiv:2403.05087}, 2024.

\bibitem[Song et~al.(2023)Song, Chen, Li, Chen, Chen, Yuan, Xu, and Geiger]{song2023nerfplayer}
Liangchen Song, Anpei Chen, Zhong Li, Zhang Chen, Lele Chen, Junsong Yuan, Yi Xu, and Andreas Geiger.
\newblock Nerfplayer: A streamable dynamic scene representation with decomposed neural radiance fields.
\newblock \emph{IEEE Transactions on Visualization and Computer Graphics}, 2023.

\bibitem[Stearns et~al.(2024)Stearns, Harley, Uy, Dubost, Tombari, Wetzstein, and Guibas]{stearns2024dgmarbles}
Colton Stearns, Adam~W. Harley, Mikaela Uy, Florian Dubost, Federico Tombari, Gordon Wetzstein, and Leonidas Guibas.
\newblock Dynamic gaussian marbles for novel view synthesis of casual monocular videos.
\newblock In \emph{ArXiv}, 2024.

\bibitem[Stich et~al.(2008)Stich, Linz, Albuquerque, and Magnor]{stich2008view}
Timo Stich, Christian Linz, Georgia Albuquerque, and Marcus Magnor.
\newblock View and time interpolation in image space.
\newblock \emph{Computer Graphics Forum}, 2008.

\bibitem[Sturm et~al.(2012)Sturm, Engelhard, Endres, Burgard, and Cremers]{sturm2012benchmark}
J{\"u}rgen Sturm, Nikolas Engelhard, Felix Endres, Wolfram Burgard, and Daniel Cremers.
\newblock A benchmark for the evaluation of rgb-d slam systems.
\newblock In \emph{2012 IEEE/RSJ international conference on intelligent robots and systems}, pages 573--580. IEEE, 2012.

\bibitem[Sumner et~al.(2007)Sumner, Schmid, and Pauly]{sumner2007embedded}
Robert~W Sumner, Johannes Schmid, and Mark Pauly.
\newblock Embedded deformation for shape manipulation.
\newblock In \emph{ACM siggraph 2007 papers}, pages 80--es. 2007.

\bibitem[Svitov et~al.(2024)Svitov, Morerio, Agapito, and Del~Bue]{svitov2024haha}
David Svitov, Pietro Morerio, Lourdes Agapito, and Alessio Del~Bue.
\newblock Haha: Highly articulated gaussian human avatars with textured mesh prior.
\newblock \emph{arXiv preprint arXiv:2404.01053}, 2024.

\bibitem[Teed and Deng(2020)]{raft}
Zachary Teed and Jia Deng.
\newblock Raft: Recurrent all-pairs field transforms for optical flow.
\newblock In \emph{Computer Vision--ECCV 2020: 16th European Conference, Glasgow, UK, August 23--28, 2020, Proceedings, Part II 16}, pages 402--419. Springer, 2020.

\bibitem[Teed and Deng(2021)]{teed2021droid}
Zachary Teed and Jia Deng.
\newblock Droid-slam: Deep visual slam for monocular, stereo, and rgb-d cameras.
\newblock \emph{Advances in neural information processing systems}, 34:\penalty0 16558--16569, 2021.

\bibitem[Teed et~al.(2024)Teed, Lipson, and Deng]{teed2024deep}
Zachary Teed, Lahav Lipson, and Jia Deng.
\newblock Deep patch visual odometry.
\newblock \emph{Advances in Neural Information Processing Systems}, 36, 2024.

\bibitem[Tian et~al.(2023)Tian, Du, and Duan]{tian2023mononerf}
Fengrui Tian, Shaoyi Du, and Yueqi Duan.
\newblock Mononerf: Learning a generalizable dynamic radiance field from monocular videos.
\newblock In \emph{Proceedings of the IEEE/CVF International Conference on Computer Vision}, pages 17903--17913, 2023.

\bibitem[Tran and Liu(2018)]{tran2018nonlinear}
Luan Tran and Xiaoming Liu.
\newblock Nonlinear 3d face morphable model.
\newblock In \emph{Proceedings of the IEEE conference on computer vision and pattern recognition}, pages 7346--7355, 2018.

\bibitem[Tretschk et~al.(2021)Tretschk, Tewari, Golyanik, Zollh{\"o}fer, Lassner, and Theobalt]{tretschk2021non}
Edgar Tretschk, Ayush Tewari, Vladislav Golyanik, Michael Zollh{\"o}fer, Christoph Lassner, and Christian Theobalt.
\newblock Non-rigid neural radiance fields: Reconstruction and novel view synthesis of a dynamic scene from monocular video.
\newblock In \emph{Proceedings of the IEEE/CVF International Conference on Computer Vision}, pages 12959--12970, 2021.

\bibitem[Wang et~al.(2021)Wang, Eckart, Lucey, and Gallo]{wang2021neural}
Chaoyang Wang, Ben Eckart, Simon Lucey, and Orazio Gallo.
\newblock Neural trajectory fields for dynamic novel view synthesis.
\newblock \emph{arXiv preprint arXiv:2105.05994}, 2021.

\bibitem[Wang et~al.(2024{\natexlab{a}})Wang, Zhuang, Siarohin, Cao, Qian, Lee, and Tulyakov]{wang2024diffusion}
Chaoyang Wang, Peiye Zhuang, Aliaksandr Siarohin, Junli Cao, Guocheng Qian, Hsin-Ying Lee, and Sergey Tulyakov.
\newblock Diffusion priors for dynamic view synthesis from monocular videos.
\newblock \emph{arXiv preprint arXiv:2401.05583}, 2024{\natexlab{a}}.

\bibitem[Wang et~al.(2024{\natexlab{b}})Wang, Ye, Gao, Austin, Li, and Kanazawa]{som2024}
Qianqian Wang, Vickie Ye, Hang Gao, Jake Austin, Zhengqi Li, and Angjoo Kanazawa.
\newblock Shape of motion: 4d reconstruction from a single video.
\newblock 2024{\natexlab{b}}.

\bibitem[Wang et~al.(2024{\natexlab{c}})Wang, Leroy, Cabon, Chidlovskii, and Revaud]{wang2024dust3r}
Shuzhe Wang, Vincent Leroy, Yohann Cabon, Boris Chidlovskii, and Jerome Revaud.
\newblock {Dust3r: Geometric 3D Vision Made Easy}.
\newblock In \emph{Proceedings of the IEEE/CVF Conference on Computer Vision and Pattern Recognition}, pages 20697--20709, 2024{\natexlab{c}}.

\bibitem[Wen et~al.(2024)Wen, Zhao, Ren, Schwing, and Wang]{wen2024gomavatar}
Jing Wen, Xiaoming Zhao, Zhongzheng Ren, Alexander~G Schwing, and Shenlong Wang.
\newblock Gomavatar: Efficient animatable human modeling from monocular video using gaussians-on-mesh.
\newblock \emph{arXiv preprint arXiv:2404.07991}, 2024.

\bibitem[Weng et~al.(2022)Weng, Curless, Srinivasan, Barron, and Kemelmacher-Shlizerman]{weng2022humannerf}
Chung-Yi Weng, Brian Curless, Pratul~P Srinivasan, Jonathan~T Barron, and Ira Kemelmacher-Shlizerman.
\newblock Humannerf: Free-viewpoint rendering of moving people from monocular video.
\newblock In \emph{Proceedings of the IEEE/CVF Conference on Computer Vision and Pattern Recognition}, pages 16210--16220, 2022.

\bibitem[Wu et~al.(2023)Wu, Yi, Fang, Xie, Zhang, Wei, Liu, Tian, and Wang]{wu20234d}
Guanjun Wu, Taoran Yi, Jiemin Fang, Lingxi Xie, Xiaopeng Zhang, Wei Wei, Wenyu Liu, Qi Tian, and Xinggang Wang.
\newblock 4d gaussian splatting for real-time dynamic scene rendering.
\newblock \emph{arXiv preprint arXiv:2310.08528}, 2023.

\bibitem[Wu et~al.(2022)Wu, Zhong, Tagliasacchi, Cole, and Oztireli]{wu2022d}
Tianhao Wu, Fangcheng Zhong, Andrea Tagliasacchi, Forrester Cole, and Cengiz Oztireli.
\newblock D2 nerf: Self-supervised decoupling of dynamic and static objects from a monocular video.
\newblock \emph{arXiv preprint arXiv:2205.15838}, 2022.

\bibitem[Xian et~al.(2021)Xian, Huang, Kopf, and Kim]{xian2021space}
Wenqi Xian, Jia-Bin Huang, Johannes Kopf, and Changil Kim.
\newblock Space-time neural irradiance fields for free-viewpoint video.
\newblock In \emph{Proceedings of the IEEE/CVF Conference on Computer Vision and Pattern Recognition}, pages 9421--9431, 2021.

\bibitem[Xiao et~al.(2024)Xiao, Wang, Zhang, Xue, Peng, Shen, and Zhou]{SpatialTracker}
Yuxi Xiao, Qianqian Wang, Shangzhan Zhang, Nan Xue, Sida Peng, Yujun Shen, and Xiaowei Zhou.
\newblock Spatialtracker: Tracking any 2d pixels in 3d space.
\newblock In \emph{Proceedings of the IEEE/CVF Conference on Computer Vision and Pattern Recognition (CVPR)}, 2024.

\bibitem[Yang et~al.(2021)Yang, Sun, Jampani, Vlasic, Cole, Chang, Ramanan, Freeman, and Liu]{yang2021lasr}
Gengshan Yang, Deqing Sun, Varun Jampani, Daniel Vlasic, Forrester Cole, Huiwen Chang, Deva Ramanan, William~T Freeman, and Ce Liu.
\newblock Lasr: Learning articulated shape reconstruction from a monocular video.
\newblock In \emph{Proceedings of the IEEE/CVF Conference on Computer Vision and Pattern Recognition}, pages 4925--4935, 2021.

\bibitem[Yang et~al.(2022)Yang, Vo, Neverova, Ramanan, Vedaldi, and Joo]{yang2022banmo}
Gengshan Yang, Minh Vo, Natalia Neverova, Deva Ramanan, Andrea Vedaldi, and Hanbyul Joo.
\newblock Banmo: Building animatable 3d neural models from many casual videos.
\newblock In \emph{Proceedings of the IEEE/CVF Conference on Computer Vision and Pattern Recognition}, pages 22247--22257, 2022.

\bibitem[Yang et~al.(2024)Yang, Kang, Huang, Xu, Feng, and Zhao]{yang2024depth}
Lihe Yang, Bingyi Kang, Zilong Huang, Xiaogang Xu, Jiashi Feng, and Hengshuang Zhao.
\newblock Depth anything: Unleashing the power of large-scale unlabeled data.
\newblock In \emph{Proceedings of the IEEE/CVF Conference on Computer Vision and Pattern Recognition}, pages 10371--10381, 2024.

\bibitem[Yang et~al.(2023{\natexlab{a}})Yang, Gao, Zhou, Jiao, Zhang, and Jin]{yang2023deformable}
Ziyi Yang, Xinyu Gao, Wen Zhou, Shaohui Jiao, Yuqing Zhang, and Xiaogang Jin.
\newblock Deformable 3d gaussians for high-fidelity monocular dynamic scene reconstruction.
\newblock \emph{arXiv preprint arXiv:2309.13101}, 2023{\natexlab{a}}.

\bibitem[Yang et~al.(2023{\natexlab{b}})Yang, Yang, Pan, Zhu, and Zhang]{yang2023real}
Zeyu Yang, Hongye Yang, Zijie Pan, Xiatian Zhu, and Li Zhang.
\newblock Real-time photorealistic dynamic scene representation and rendering with 4d gaussian splatting.
\newblock \emph{arXiv preprint arXiv:2310.10642}, 2023{\natexlab{b}}.

\bibitem[Yoon et~al.(2020)Yoon, Kim, Gallo, Park, and Kautz]{yoon2020novel}
Jae~Shin Yoon, Kihwan Kim, Orazio Gallo, Hyun~Soo Park, and Jan Kautz.
\newblock Novel view synthesis of dynamic scenes with globally coherent depths from a monocular camera.
\newblock In \emph{Proceedings of the IEEE/CVF Conference on Computer Vision and Pattern Recognition}, pages 5336--5345, 2020.

\bibitem[You and Hou(2023)]{you2023decoupling}
Meng You and Junhui Hou.
\newblock Decoupling dynamic monocular videos for dynamic view synthesis.
\newblock \emph{arXiv preprint arXiv:2304.01716}, 2023.

\bibitem[Yu et~al.(2024)Yu, Sattler, and Geiger]{gof}
Zehao Yu, Torsten Sattler, and Andreas Geiger.
\newblock Gaussian opacity fields: Efficient and compact surface reconstruction in unbounded scenes.
\newblock \emph{arXiv preprint arXiv:2404.10772}, 2024.

\bibitem[Zhang et~al.(2024)Zhang, Herrmann, Hur, Jampani, Darrell, Cole, Sun, and Yang]{zhang2024monst3r}
Junyi Zhang, Charles Herrmann, Junhwa Hur, Varun Jampani, Trevor Darrell, Forrester Cole, Deqing Sun, and Ming-Hsuan Yang.
\newblock Monst3r: A simple approach for estimating geometry in the presence of motion.
\newblock \emph{arXiv preprint arxiv:2410.03825}, 2024.

\bibitem[Zhang et~al.(2022)Zhang, Cole, Li, Rubinstein, Snavely, and Freeman]{zhang2022structure}
Zhoutong Zhang, Forrester Cole, Zhengqi Li, Michael Rubinstein, Noah Snavely, and William~T Freeman.
\newblock Structure and motion from casual videos.
\newblock In \emph{European Conference on Computer Vision}, pages 20--37. Springer, 2022.

\bibitem[Zhao et~al.(2022)Zhao, Liu, Guo, Wang, and Liu]{zhao2022particlesfm}
Wang Zhao, Shaohui Liu, Hengkai Guo, Wenping Wang, and Yong-Jin Liu.
\newblock {ParticleSfM: Exploiting Dense Point Trajectories for Localizing Moving Cameras in the Wild}.
\newblock In \emph{European conference on computer vision (ECCV)}, 2022.

\bibitem[Zhao et~al.(2024)Zhao, Colburn, Ma, Bautista, Susskind, and Schwing]{zhao2024pseudogeneralized}
Xiaoming Zhao, Alex Colburn, Fangchang Ma, Miguel~Angel Bautista, Joshua~M. Susskind, and Alexander~G. Schwing.
\newblock Pseudo-generalized dynamic view synthesis from a video, 2024.

\bibitem[Zhou et~al.(2024)Zhou, Zhong, Shin, Lu, Yang, Markham, and Trigoni]{zhou2024dynpoint}
Kaichen Zhou, Jia-Xing Zhong, Sangyun Shin, Kai Lu, Yiyuan Yang, Andrew Markham, and Niki Trigoni.
\newblock Dynpoint: Dynamic neural point for view synthesis.
\newblock \emph{Advances in Neural Information Processing Systems}, 36, 2024.

\bibitem[Zitnick et~al.(2004)Zitnick, Kang, Uyttendaele, Winder, and Szeliski]{zitnick2004high}
C.~Lawrence Zitnick, Sing~Bing Kang, Matthew Uyttendaele, Simon Winder, and Richard Szeliski.
\newblock High-quality video view interpolation using a layered representation.
\newblock \emph{ACM Transactions on Graphics (TOG)}, 2004.

\bibitem[Zollh{\"o}fer et~al.(2014)Zollh{\"o}fer, Nie{\ss}ner, Izadi, Rehmann, Zach, Fisher, Wu, Fitzgibbon, Loop, Theobalt, et~al.]{zollhofer2014real}
Michael Zollh{\"o}fer, Matthias Nie{\ss}ner, Shahram Izadi, Christoph Rehmann, Christopher Zach, Matthew Fisher, Chenglei Wu, Andrew Fitzgibbon, Charles Loop, Christian Theobalt, et~al.
\newblock Real-time non-rigid reconstruction using an rgb-d camera.
\newblock \emph{ACM Transactions on Graphics (ToG)}, 33\penalty0 (4):\penalty0 1--12, 2014.

\end{thebibliography}
}

\end{document}